\PassOptionsToPackage{numbers}{natbib}
\PassOptionsToPackage{sort}{natbib}

\documentclass{article}

\usepackage{geometry}
\usepackage[numbers]{natbib}
\usepackage{fullpage}
\date{}

\usepackage[utf8]{inputenc} 
\usepackage[T1]{fontenc}    
\usepackage{lmodern}
\usepackage{hyperref}       
\usepackage{url}            
\usepackage{booktabs}       
\usepackage{amsfonts}       
\usepackage{nicefrac}       
\usepackage{microtype}      
\usepackage{epsfig}
\usepackage{graphicx}
\usepackage{amsmath}
\usepackage{amssymb}
\usepackage{xcolor}
\usepackage{mathtools}
\usepackage{lipsum}
\usepackage{wrapfig}
\usepackage[ruled,vlined]{algorithm2e}

\usepackage{colortbl}
\usepackage{nccmath}
\usepackage{bm}

\usepackage{array}
\newcolumntype{C}[1]{>{\centering\let\newline\\\arraybackslash\hspace{0pt}}m{#1}}

\usepackage{xcolor}

\makeatletter
\newcommand{\ccell}[3][]{%
	\kern-\fboxsep
	\if\relax\detokenize{#1}\relax
	\expandafter\@firstoftwo
	\else
	\expandafter\@secondoftwo
	\fi
	{\colorbox{#2}}%
	{\colorbox[#1]{#2}}%
	{#3}\kern-\fboxsep
}
\makeatother
\definecolor{cellgray}{gray}{0.8}
\AtBeginDocument{\setlength{\cmidrulekern}{0.3em}}

\title{Lookahead Adversarial Learning for Near Real-Time Semantic Segmentation}

\author{Hadi Jamali-Rad \\ \small Shell Global Solutions International B.V. \\ \footnotesize \texttt{\href{mailto:hadi.jamali-rad@shell.com}{hadi.jamali-rad@shell.com}} \and
Attila Szab\'{o} \\ \small Shell Global Solutions International B.V. \\ \footnotesize \texttt{\href{mailto:attila.szabo@shell.com}{attila.szabo@shell.com}}}

\begin{document}

\maketitle
\begin{abstract}
Semantic segmentation is one of the most fundamental problems in computer vision with significant impact on a wide variety of applications. Adversarial learning is shown to be an effective approach for improving semantic segmentation quality by enforcing higher-level pixel correlations and structural information. However, state-of-the-art semantic segmentation models cannot be easily plugged into an adversarial setting because they are not designed to accommodate convergence and stability issues in adversarial networks. We bridge this gap by building a conditional adversarial network with a state-of-the-art segmentation model (DeepLabv3+) at its core. To battle the stability issues, we introduce a novel lookahead adversarial learning (LoAd) approach with an embedded label map aggregation module. We focus on semantic segmentation models that run fast at inference for near real-time field applications. Through extensive experimentation, we demonstrate that the proposed solution can alleviate divergence issues in an adversarial semantic segmentation setting and results in considerable performance improvements ($+5\%$ in some classes) on the baseline for three standard datasets.\footnote{The paper is under consideration at Computer Vision and Image Understanding.}
       
\end{abstract}

\section{Introduction}
\label{sec:intro}
\vspace{-0.5cm}
Semantic segmentation is a challenging task in computer vision. It is a pivotal step towards content-based image analysis and scene understanding as it empowers machines to distinguish between different regions of an image based on its semantic context. To this aim, semantic segmentation models are trained to assign semantic labels to each and every pixel of an image as well as to cluster them into groups. Semantic segmentation has received an upsurge of attention recently owing to its wide variety of applications in medical imaging \cite{ronneberger2015u, rezaei2017conditional}, autonomous driving \cite{menze2015object, cordts2016cityscapes}, satellite image processing \cite{volpi2015semantic, henry2018road}, and robotics \cite{geiger2013vision, shvets2018automatic}, to name a few. Early segmentation methodologies are mostly developed with clustering algorithms at their core and somehow trying to incorporate contour, edge, and structural information. Examples of such algorithms are active contours \cite{kass1988snakes}, region-growing \cite{nock2004statistical}, conditional random fields (CRFs) \cite{plath2009multi}, and sparse reconstruction based methods \cite{minaee2019admm}. Recent advances in deep learning and convolutional neural networks (CNNs) revolutionized this field resulting in state-of-the-art image segmentation algorithms such as FCN \cite{long2015fully}, U-Net \cite{ronneberger2015u}, PSPNet \cite{zhao2017pyramid}, EncNet \cite{zhang2018context}, Exfuse \cite{zhang2018exfuse}, DeepLabv3+ \cite{chen2018encoder}, PS and Panoptic DeepLab \cite{kirillov2019panoptic, cheng2019panoptic},  HRNet \cite{wang2020deep} and many other elegant architectures that considerably outperformed the traditional signal processing based methods addressing the same challenge. 

Majority of these deep learning based methods formulate semantic segmentation as a classification problem where cross entropy (CE) with pixel independence assumption is employed as the optimization loss function. However, in practice, adjacent pixels of an image are highly correlated. These methods implicitly assume that correlation among pixels would be learned as receptive field of CNNs increases going deeper with convolutions. Recent studies challenge this assumption and argue that overlooking pixel correlations explicitly can lead to performance degradation especially with regards to capturing structural information embedded in target classes. These studies have proposed different approaches to capture pixel inter-dependencies. For instance, CRFs can be used to model pixel relationships and enforce label consistency between pixels \cite{liu2017deep, chen2017deeplab, shen2017semantic, liu2017deep}. However, CRFs are known to be time-consuming at inference and sensitive to variations in visual appearance. An alternative approach is extracting pixel affinity information from images and fusing them back to predicted label maps \cite{ke2018adaptive}; this comes at the cost of extra model branches and larger memory requirements. More recent studies have proposed changing the perspective and using different loss functions that encode the mutual information or structural similarity among nearby pixels in a regional fashion \cite{zhao2019region, zhao2019correlation} and have shown improvements. However, these losses can be derived in a sub-optimal manner by making major simplifying assumptions and by considering a small patch of pixels.  

Another avenue that has been explored to enforce structure in segmentation is employing adversarial learning \cite{luc2016semantic, souly2017semi, xue2018segan, hung2018adversarial}. In this setup, a segmentor-discriminator pair compete to outperform each other in creating realistic label maps and distinguishing them from ground truth ones. We think a conditional adversarial approach similar to \cite{luc2016semantic, souly2017semi} has the capacity to capture these pixel inter-dependencies and correlations in a more general (and not only local) fashion when compared to methods proposed in \cite{zhao2019region, zhao2019correlation}. In medical imaging, there are studies proposing (generative) adversarial networks for image segmentation \cite{imran2020progressive, chen2018semantic}; however, their aim is tackling domain shift in semi-supervised settings and they mostly employ U-Net type segmentors given their solid performance in this context. On the other hand, plugging state-of-the-art semantic segmentation models in an adversarial setting is prone to the well-known divergence and mode collapse issues \cite{arjovsky2017wasserstein, goodfellow2014explaining}. Specific architecture designs for generator and discriminator networks can help to stabilize the setup, but at the cost of limiting the application domain of adversarial networks. When it comes to semantic segmentation, we are bound to architectures specifically designed to excel in doing so. When we tried to establish an adversarial network with DeepLabv3+ \cite{chen2018encoder} as segmentor, we could not manage to stabilize the network regardless of remedies we pulled in. Bridging the gap between employing the state-of-the-art semantic segmentation models in adversarial settings and helping to stabilize them is the core idea of the proposed \emph{lookahead adversarial learning} (LoAd) approach. Notably, we focus on models that run \emph{fast} at inference time for \emph{near real-time} field applications. That is why we opt for DeepLabv3+ base models that offer speed (\emph{no multi-scaling (MS), no CRFs}) and performance at the same time. Nonetheless, in our ablation studies, we also demonstrate the impact of applying time-consuming post-processing steps such as MS on the performance of proposed solution.  

The proposed solution (LoAd) takes inspiration from the ``lookahead optimizer'' \cite{zhang2019lookahead} and allows the semantic segmentation adversarial network to go ahead and actually diverge to some extent, then inspired by DAGGER in imitation learning \cite{ross2011reduction} aggregates the degraded label maps, and steps backward to use these new sets of information and avoid further divergence. Notably, our label map aggregation strategy is different from collecting adversarial examples to retrain and stabilize generative adversarial networks (GANs) \cite{goodfellow2014explaining, liu2019rob}. We demonstrate that LoAd can alleviate divergence issues of adversarial training in a semantic segmentation setting leading to improvement in mean-intersection-over-union (mIoU) over the baseline DeepLabv3+ on Pascal VOC 2012, Cityscapes and Stanford Background datasets. We also show that in some classes $+5\%$ improvement in IoU beyond the baseline is achieved which is quite significant. We then qualitatively demonstrate that our solution is boosting the baseline in overall segmentation performance in tackling class swap/confusion, as well as in better understanding of structure and continuity of the target classes.

The main contributions of this paper can be summarized as follows: a) we propose a lookahead adversarial learning method (LoAd) for semantic segmentation that helps alleviate stability issues in such settings, b) LoAd runs as fast as the baselines methods upon which it is applied, i.e., no extra delay at inference time, making it a great choice for near real-time field applications, c) besides avoiding class confusion, LoAd improves the performance of the baseline in creating structurally more consistent label maps with significant performance boost in some classes. Enhancing state-of-the-art segmentation models that can be trained on \emph{commodity GPUs} and run \emph{fast at inference} time (no CRFs, no multi-scaling) is the motive of this work.
\section{Conditional Adversarial Training for Semantic Segmentation}
\label{sec:adversarial}
\vspace{-0.4cm}
Let $\mathcal{D}_t =\{({\bf X},{\bf Y})_{1},...,({\bf X},{\bf Y})_{M}\}$ be the training dataset containing $M$ samples with $\mathcal{X}_t=\{{\bf X}|({\bf X},{\bf Y}) \in \mathcal{D}_t \}$ and $\mathcal{Y}_t=\{{\bf Y}|({\bf X},{\bf Y}) \in \mathcal{D}_t \}$ respectively denoting the set of images and their corresponding label maps\footnote{From now on, we use ``map'' and ``label map'' interchangeably.}. Here, ${\bf X}$ is of size $H \times W \times 3$ for RGB images with a total of $H \times W = N$ pixels. The corresponding label map ${\bf Y}$ is of size $H \times W$ with elements in $\mathcal{K}=\{1,\dotsc,K\}$ where $K$ is the number of classes in the segmentation task. An adaptation of conditional generative adversarial networks (CGANs) \cite {goodfellow2014generative, mirza2014conditional, isola2017image} for semantic segmentation would not require stochastic behavior in generating semantic label maps but aims at creating the most probable map ${\bf Y}$ per input image ${\bf X}$. So, we solve a two-player min-max game to estimate $P({\bf Y}|{\bf X})$
\begin{equation}
\label{eq:advLoss}
\min_{G} \max_{D} \mathcal{L}(G,D) = \mathbb{E}_{{\bf Y} \sim P_{\mathcal{D}_t}({\bf Y})}\big[\log{(D({\bf Y}|{\bf X}))}\big] + \mathbb{E}_{{\bf Y} \sim P_{g}({\bf Y})}\big[\log{(1-D({\bf Y}|{\bf X}))}\big], \\
\end{equation} 
where $\mathcal{L}(.)$ denotes the loss function, $G$ denotes the generator (more specifically a segmentor) parameterized with $\boldsymbol{\theta}_g$, and $D$ stands for the discriminator parameterized with $\boldsymbol{\theta}_d$. Typically, both $G$ and $D$ are CNN's. This can be further simplified within a binary classification setting where the discriminator is to decide whether a sample label map is ground truth (${\bf Y} \sim P_{\mathcal{D}_t}$) or generated (${\bf Y} \sim P_{g}$) by the segmentor. Considering binary cross entropy (CE), we arrive at the following adversarial loss    
\begin{align}
\label{eq:advLoss_2}
\mathcal{L}_a (\boldsymbol{\theta}_g, \boldsymbol{\theta}_d) = \sum_{m=1}^M \big[\log\big(D({\bf Y}_m|{\bf X}_m)\big) + \log\big(1 - D(\hat{{\bf Y}}_m|{\bf X}_m)\big)\big], 
\end{align}
where $\hat{{\bf Y}}$ denotes the generated map. $\mathcal{L}_a$ should be minimized w.r.t. $\boldsymbol{\theta}_g$ and maximized w.r.t $\boldsymbol{\theta}_d$. Several interesting studies such as \cite{luc2016semantic, souly2017semi} suggest applying a hybrid loss combining the conditional adversarial loss in \eqref{eq:advLoss_2} with a regularization or CE pixel-wise term, sometimes in a slightly different context such as weakly supervised GAN settings \cite{souly2017semi}. The closest approach to our line of thought is the pioneering work in \cite{luc2016semantic} where the following hybrid loss $\mathcal{L}_h = \mathcal{L}_p + \lambda \, \mathcal{L}_a$ is proposed
\begin{align}
\label{eq:hybLoss_2}
\mathcal{L}_h = \sum_{m=1}^M\textup{CE}({\bf Y}_m, \hat{{\bf Y}}_m) + \lambda \, \sum_{m=1}^M \log\big(D({\bf Y}_m|{\bf X}_m)\big) + \lambda \, \sum_{m=1}^M \log\big(1 - D(\hat{{\bf Y}}_m|{\bf X}_m)\big), 
\end{align} 
%
where the pixel-wise loss $\mathcal{L}_p$ is computed using a multi-class CE between the $1$-hot encoded versions of the original label map ${\bf Y}$ and the inferred one $\hat{{\bf Y}}$ using $- \sum_{i=1}^N \sum_{c = 1}^{K} y_{i,c} \, \log(\hat{y}_{i,c})$, with $y_i$ denoting the $i$th element of ${\bf Y}$. Obviously, only the second and the third terms in \eqref{eq:hybLoss_2} are relevant when training the discriminator. When training the generator, \cite{luc2016semantic} proposes to keep the first and the third terms. Next, a standard gradient decent ascent (GDA) \cite{lin2019gradient} is applied to the two-player min-max game. 

We decided to take a different approach for two reasons. First, interesting findings are recently reported in \cite{nouiehed2019solving, ostrovskii2020efficient} regarding the optimization of $\min_{x \in {\bf X}} \max_{y \in {\bf Y}} F (x, y)$ problems where $F(x, y)$ is concave in $y$ and non-convex in $x$, which relates to our problem in that the loss is concave in $\boldsymbol{\theta}_d$ and non-convex in $\boldsymbol{\theta_g}$ in typical high dimensional settings with CNNs. Therefore, following the propositions in \cite{nouiehed2019solving, ostrovskii2020efficient}, and (to our understanding) in contrast to \cite{luc2016semantic}, we avoid an alternating GDA in optimizing the generator and discriminator networks. Instead, in every ``cycle'' of the proposed adversarial approach (Algorithm~\ref{alg:lookahead_simplified}), we keep training the discriminator with a dynamically updated dataset according to the proposed label map aggregation module (Algorithm~\ref{alg:buffer}) to reach sufficient accuracy before switching back to training the generator. 

Second, another angle that distinguishes us from \cite{luc2016semantic} is in how and when to incorporate the pixel-wise CE loss. We approach the problem in two stages as follows. Stage 1: if not pre-trained on $\mathcal{D}_t$, we first train the segmentation network using only CE pixel-wise loss up to a reasonable performance (without hard constraints). Stage 2: we then activate the adversarial loss on top the CE loss and run (LoAd) as described in Algorithm~\ref{alg:lookahead_simplified} to boost the performance. At this stage, when training discriminator both the second and the third terms of \eqref{eq:hybLoss_2} will be active, and when training the segmentation network the first and the third terms will be used where we follow the suggestion of \cite{luc2016semantic, goodfellow2014generative} and replace $+\lambda \sum_{m=1}^M \log(1 - D(\hat{{\bf Y}}_m|{\bf X}_m))$ with $-\lambda \sum_{m=1}^M \log(D(\hat{{\bf Y}}_m|{\bf X}_m))$ to maximize the probability of $\hat{{\bf Y}}_m$ being the true segmentation label map of ${\bf X}_m$.

\section{Lookahead Adversarial Learning (LoAd)}
\label{sec:lookaeahd}
\vspace{-0.4cm}
{\bf The idea.} Robustness and divergence issues of GANs are not secret to anyone \cite{arjovsky2017wasserstein, goodfellow2014explaining, liu2019rob, roth2017stabilizing, salimans2016improved}. Generally speaking, adversarial networks manifest the same difficulties and we had to tackle them in our semantic segmentation setup. Here is the idea behind lookahead adversarial learning (LoAd) for semantic segmentation in a nutshell. We take inspiration from ``lookahead optimizer'' \cite{zhang2019lookahead} and allow the adversarial network to \emph{go ahead} and actually diverge (to some extent) helping us to gain new insights and construct new datasets of label maps from these divergent (or degraded) models. Inspired by the idea of DAGGER \cite{ross2011reduction}, we aggregate these new datasets and use them for retraining the discriminator at the end of every cycle of LoAd. Next, we go back to where the divergence started (similar to ``$1$ step back'' in lookahead optimizer) to improve our next predictions and avoid further divergence. These new datasets are not designed or generated adversarial examples but sequentially degraded label maps. Note the importance of label maps in this context. When we descend towards divergence not only our performance metric goes down (mIoU, a single score) but also destructive impacts on the generated maps provides us with new sets of information that we exploit. 

\begin{figure}[b]
	\centering
	\includegraphics[width=0.8\textwidth]{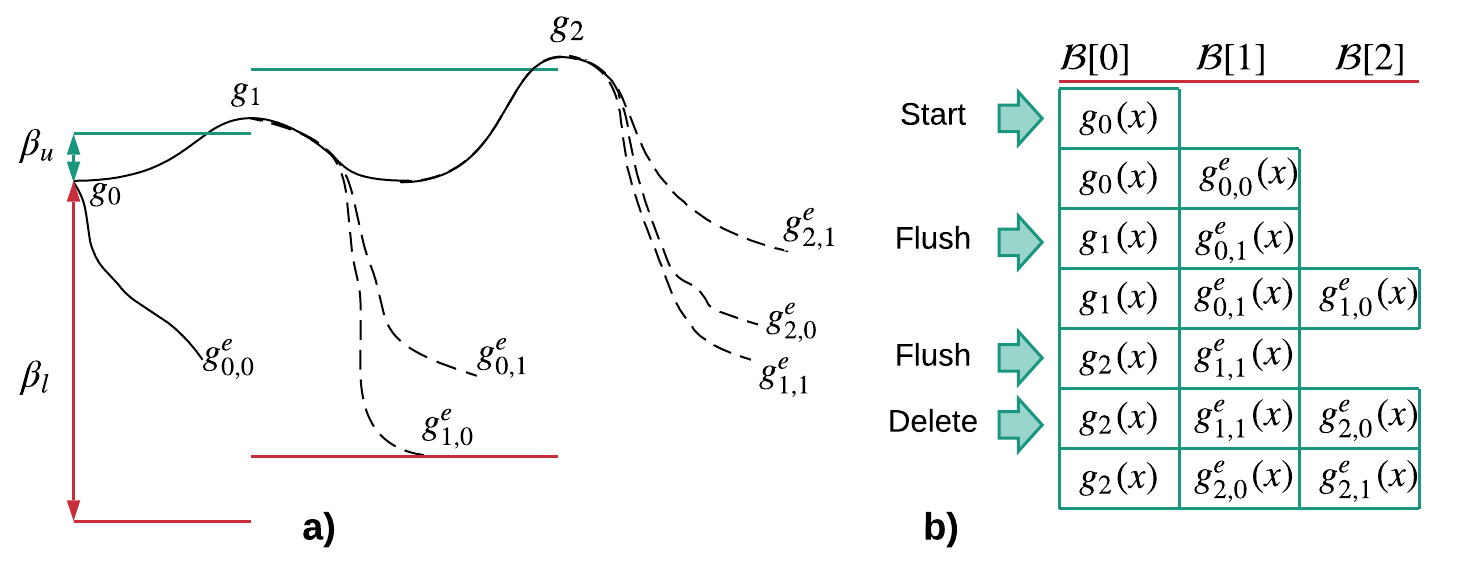}
	\vspace{-0.3cm}
	\caption{a) hypothetical convergence graph, b) corresponding label map aggregation buffer.}
	\vspace{-0.4cm}
	\label{fig:maskaggregation}
	\vspace{-0.2cm}
\end{figure}
{\bf The mechanics.} Algorithm~\ref{alg:buffer} describes the map aggregation module of LoAd and Algorithm~\ref{alg:lookahead_simplified} provides a pseudo code level description of LoAd itself. Let us assume a starting model $g^{s} = g_0$ (e.g., the ending model after Stage 1 training as explained in Section~\ref{sec:adversarial}). First, we evaluate the model on a subset of validation data (a hold-out set) to understand our current mIoU (denoted as $\mu$ in Algorithm~\ref{alg:lookahead_simplified}). This serves as both starting and current best mIoU ($\mu^{s}$ and $\mu^{*}$, respectively). We can already train our discriminator for the first time using $\mathcal{D}_t \cup g_0(\mathcal{X}_t)$, a set composed of full training data (images and maps) plus a set of generated (fake) maps. With this, we have initialized our label map aggregation buffer with $\mathcal{B} = \mathcal{B}[0] = g_0(\mathcal{X}_t)$ in Algorithm~\ref{alg:buffer}. Back to Algorithm~\ref{alg:lookahead_simplified}, we then continue training adversarial until one of the following two criteria is met: a) patience iteration counter $\gamma$ reaches its maximum $\Gamma$, alerting us that it is enough looking ahead, b) we diverge (in mIoU sense) reaching a pre-defined lower-bound ($\mu^{s} - \beta_l$) w.r.t. to the starting mIoU $\mu^{s}$. If any of the two criteria are met, the cycle is finished, and we pick the last model of the cycle denoted by $g^{e}$. Throughout each cycle we also seek for an updated model offering a mIoU better than the staring one, and if such a new peak model $g^{*}$ (above an upper-bound $\mu^{s} + \beta_u$) is found, the cycle would be returning two models, the best model of the cycle $g^{*}$ besides its ending model $g^{e}$. 

%
\begin{algorithm}[!ht]
	\SetKwInput{Return}{Return}
	\SetKwInput{Require}{Require}
	\SetAlgoLined
	\DontPrintSemicolon
	\SetNoFillComment
	\Require{peak/ending models, image set $(g^{p}, g^{e}, \mathcal{X})$}
	\eIf{$g^{p} \neq 0$}{
		\texttt{\textup{Flush}}$(\mathcal{B})$\;
		$\mathcal{B}[0] \gets g^{p}(\mathcal{X})$\;
		$\mathcal{B}[1] \gets g^{e}(\mathcal{X})$}{
		\If{\texttt{\textup{Size}}$(\mathcal{B}) = $ \textup{max buffer size} $(B_{max})$}{
			\texttt{\textup{Delete}}$(\mathcal{B}[1])$\;
		}{$\mathcal{B}[\textup{end}] \gets g^{e}(\mathcal{X}) $\;}
	}
	\Return{$\mathcal{B}$}
	\caption{Map Aggregation Buffer}\label{alg:buffer}
\end{algorithm}
\begin{algorithm}[ht!]
	\SetKwInOut{Init}{Initialize}
	\SetAlgoLined
	\DontPrintSemicolon
	\SetNoFillComment
	\Init{$\psi = 0$, $g^{s} = g_{0}$, $\mathcal{B} = g^{s}(\mathcal{X}_t)$}
	\KwIn{\textup{maximum cycles:}\,$\Psi$, \textup{maximum patience:}\,$\Gamma$, $\beta_l$, $\beta_u$}
	$\mu^{s}, \mu^{*}, \mu \gets \textup{evaluate mIoU}$ \;
	\texttt{\textup{Train Discriminator}}$(\mathcal{D}_t \cup \mathcal{B})$\;
	\While{$\psi < \Psi$}{ 
		\textup{start a divergence patience counter:} $\gamma \gets 0$\;
		\While{$\mu^{s} - \beta_l < \mu$ \textup{and} $\gamma < \Gamma$}{
			\textup{update model:} $g \gets \texttt{\textup{Train Adversarial}}$\;
			$\mu \gets \textup{evaluate mIoU}$ \;
			$\mu^{*} \gets$ \textup{best} $\mu > \mu^{s} + \beta_{u}$\; 
			\textup{update best model:} $g^{*} \gets g$ \;
			$\gamma \gets \gamma + 1$
		}
		$g^{e} \gets \textup{keep the last model of the cycle}$\;
		\eIf{\textup{best model better than start}}{
			\textup{set best model as start model:} $g^{s} \gets g^{*}$\;
			\textup{reset cycle counter} $\psi \gets 0$\;
			$\mathcal{B} \gets$ \texttt{\textup{MapAggregation}}$(g^{*}, g^{e}, \mathcal{X})$ \;
		}{
			$\mathcal{B} \gets$ \texttt{\textup{MapAggregation}}$(0, g^{e}, \mathcal{X})$ \;
			\textup{start a new cycle $\psi \gets \psi + 1$}\;
		}
		\texttt{\textup{Train Discriminator}}$(\mathcal{D}_t \cup \mathcal{B})$\;
	}
	\caption{LoAd for Semantic Segmentation}\label{alg:lookahead_simplified}
\end{algorithm}

Per cycle one or both of these models $\{g^{*}, g^{e}\}$ would be passed to our map aggregation algorithm to generate new ``fake'' label maps which will be aggregated in $\mathcal{B}$. This dynamically updated dataset in $\mathcal{B}$ concatenated with $\mathcal{D}_t$ will then be used to retrain the discriminator before the next cycle starts. Owing to this label map aggregation and following retraining of the discriminator, we continually improve our adversarial model to avoid further divergences. At the end of each cycle, we go back and restart training adversarial from the newly found peak $g^*$ or the old starting point $g^s$. Lastly, if we do more than $\Psi$ cycles from a starting model $g^s$ and a new peak is not found to replace it, the algorithm fully stops and returns the overall best model. Notably, the divergence patience counter $\gamma$ is in practice updated in a dynamic manner (together with an auxiliary ``peak finder'' counter) to avoid an upward trend being stopped prematurely. Interested reader is referred to more details in the supplementary material.

{\bf The recap.} To make this crystal clear, we use a hypothetical convergence graph in Fig.~\ref{fig:maskaggregation} a) and corresponding dynamically updated map aggregation buffer depicted in Fig.~\ref{fig:maskaggregation} b) to walk you through what LoAd does in action. As can be seen, starting from $g_0$, the first adversarial cycle immediately descends towards divergence ending with $g^{e}_{0,0}$. We denote the $j$th cycle spawned from the $i$th peak with $g_{i,j}$. Note that $g^{e}_{0,0}$ does not descend by $\beta_l$, and thus, we are assuming that the cycle is ended due to reaching patience limit of $\Gamma$ propagations (or iterations) as described in Algorithm~\ref{alg:lookahead_simplified}. This cycle also did not introduce a new peak better than $g_0$. Thus, only $g^{e}_{0,0} (\mathcal{X})$ will be added as a new set to the buffer $\mathcal{B}$. This is where we go back and restart adversarial training from $g_{0}$, but this time with a retrained discriminator. As in the figure, this helps to ascend towards $g_1$ after which we diverge again in the second cycle. So, the second cycle returns a new peak $g^{*} = g_1$ as well as the ending model $g^{e} = g^{e}_{0,1}$ for map aggregation. Since a new peak is found (better than $g_{0}$), we \texttt{Flush} the buffer filling it with $\mathcal{B} = [g_{1}(\mathcal{X})\,|\,g^{e}_{0,1}(\mathcal{X})]$ as shown in Fig.~\ref{fig:maskaggregation} b). Per pseudo code in Algorithm~\ref{alg:buffer}, any cycle that only returns an ending model (an no new peak) would result in creating a new label map set added to the end of the buffer unless the buffer is full; i.e., it already contains $B_{max}$ label map sets. In that case, we first \texttt{Delete} the label map set corresponding to the oldest ending model and then the new label map set is added to the end of $\mathcal{B}$ as described in Algorithm~\ref{alg:buffer}. An example of this scenario in our hypothetical setup is where the set corresponding to $g^{e}_{1,1}$ is deleted in favor of the newcomer set corresponding to $g^{e}_{2,1}$. 

\section{Experimental Setup}
\label{sec:exp}
\begin{figure}[t]
	\centering
	\includegraphics[width=0.7\textwidth]{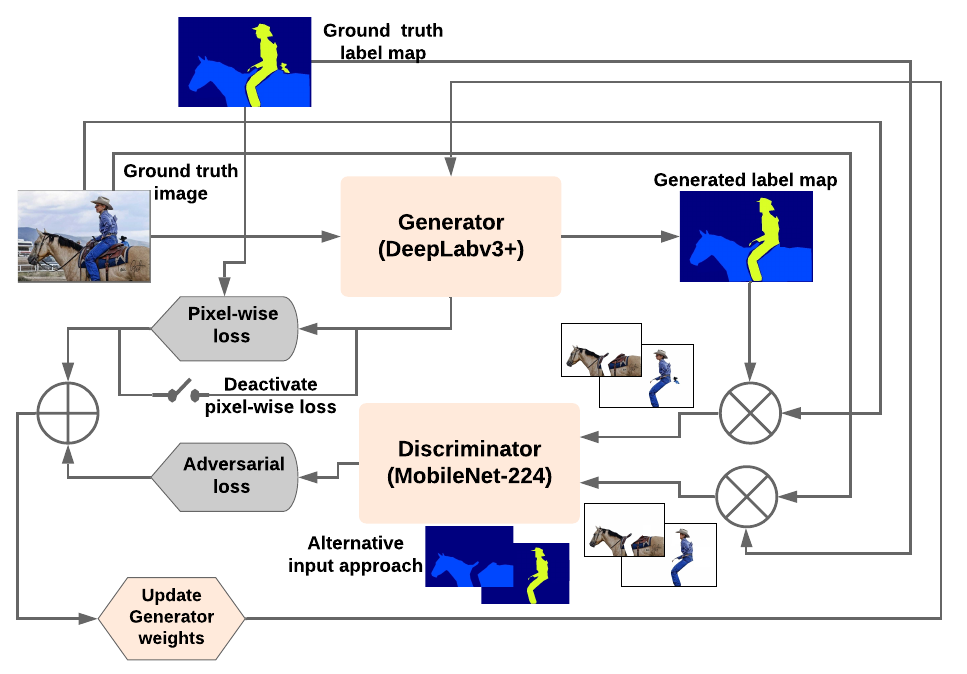}
	\vspace{-0.3cm}
	\caption{Adversarial network architecture.}
	\label{fig:network}
	\vspace{-0.3cm}
\end{figure}

\vspace{-0.4cm}
\textbf{Network architecture.} Our experimental network architecture is shown in Fig.~\ref{fig:network}. As can be seen, we opted for DeepLabv3+ with different backbones (modified Xception-$65$ \cite{chollet2017xception} and MobileNetv2 \cite{chen2018encoder}), bearing in mind that DeepLabv3+ might not be the easiest model to simply plug into an adversarial settings. We chose Mobilenet-$224$ \cite{howard2017mobilenets} as our discriminator. The figure shows our discriminator training policy being conditional on the input image split into different classes (using ground truth and generated label maps) and stacked into the input channels of the discriminator. We also considered a few other possibilities such as feeding the discriminator with split label maps themselves, which led to performance degradation. Our trainings are run separately on standard Nvidia P100 Tesla nodes offered on Microsoft Azure each with $16$\,GB of memory. We focus on models that can be trained and run on standard and \emph{commodity GPU nodes}, as many researchers do not have access to high-end TPUs. On the same note, and to re-emphasize, we are particularly interested in models that \emph{run fast at inference} time for near real-time field applications. That is why we selected DeepLabv3+ base models that offer speed (\emph{no MS}, \emph{no CRFs}) and performance at the same time.
%

\textbf{Adversarial training.} Per cycle in Algorithm~\ref{alg:lookahead_simplified}, we train MobileNet (discriminator) until sufficient performance is reached based on an early-stopping criterion evaluated on a hold-out set explained in the following. For training the discriminator, we used a batch size of $16$, and set $\alpha = 1$ with a dropout rate of $0.01$ \cite{howard2017mobilenets}. We used Adagrad as optimizer with a learning rate $lr_d = 0.01$. Because of the label map aggregation, the loss function of the discriminator (binary CE) is weighted to account for the variable number of generated label maps in the map aggregation buffer. For adversarial training of the generator, we used a batch size of $5$ due to the memory limitation of the GPU nodes. The adversarial learning rate was set to $lr_a = 5e-6$, and we trained using a momentum of $0.95$. Adversarial training is conducted according to LoAd (in Algorithm~\ref{alg:lookahead_simplified}) with the following default hyperparameters: $\beta_u = 0.001$ and $\beta_l = 0.05$ corresponding to $0.1\%$ improvement and $5\%$ drop in mIoU in a cycle, respectively; patience counter maximum is set to $\Gamma = 50$, and maximum number of cycles allowed is set to $\Psi = 50$; maximum buffer set size is $B_{max} = 3$.

\textbf{PASCAL VOC 2012 dataset.} PASCAL VOC 2012 dataset \cite{everingham2015pascal} is one of the most popular datasets for semantic segmentation with $20$ foreground object classes and $1$ background class. It contains $1464$ train, $1449$ validation, and $1456$ test pixel-level annotated images. For the experiments on this dataset, we started from DeepLabv3+ checkpoints pre-trained on PASCAL VOC 2012 with output stride of $16$ achieving mIoU $=82.2\%$ (Xception-$65$ backbone) and mIoU $=75.32\%$ (MobileNetv2 backbone) on the validation set; see, Table $5$ in \cite{chen2018encoder} and their model zoo on Github\footnote{\url{https://github.com/tensorflow/models/blob/master/research/deeplab/g3doc/model_zoo.md}}. We then applied LoAd for adversarial training. The images have a shape of at most $512 \times 512$ pixels, so we used a crop size of $513 \times 513$ following the recommendations in \cite{chen2018encoder}.

\textbf{Cityscapes dataset.} Cityscapes is one of the most commonly used large-scale datasets for semantic segmentation \cite{cordts2016cityscapes} with $19$ main target classes used for evaluation. It contains $2975$ train, $500$ validation, and $1525$ test high quality pixel-level annotated images (a total of $5000$ ``fine annotation'' images) used in our experimentation. For the experiments on this dataset, we started from DeepLabv3+ checkpoint pre-trained on Cityscapes  with output stride of $16$ achieving mIoU $=70.71\%$ on the validation set with MobileNetv2 backbone; see their model zoo on Github referred earlier. Notably, due to considerably larger input image size ($1024 \times 2048$), using Xception backbones would enforce a very small batch size for typical GPU nodes, rendering training meaningless. We applied LoAd for adversarial training with a crop size of $513 \times 513$. 

\textbf{Stanford Background dataset.} The Stanford Background dataset \cite{gould2009decomposing} contains $8$ classes of scene elements. It has $715$ pixel-level annotated images which we have split into $400$ for training, $172$ for validation and $143$ for testing. We chose this dataset to explore what the impact on smaller datasets (from sample size perspective) would be. For this dataset, there was no pre-trained DeepLabv3+ model available and we followed our two stage approach as explained in Section~\ref{sec:adversarial}. We used DeepLabv3+ checkpoint from Cityscapes achieving mIoU of $78.79\%$ (see Table $7$ in \cite{chen2018encoder}) and trained it on Stanford Background using only pixel-wise CE loss. We set batch size to $7$ and trained the model with a weighted CE loss for $4$ epochs then switched to the original CE loss (all weights set to $1$) and trained for another $8$ epochs reaching mIoU $=74.33\%$. Weighted CE loss is calculated by multiplying the loss by the average number of pixels per class computed over the entire training dataset and dividing it by the number of pixel per class in the image. This model is then used as the starting point for applying and evaluating LoAd. The learning rate for the pixel-wise training was set to $lr_p = 1e-4$ and kept constant throughout the training following \cite{chen2018encoder}. We did not use multi-scale training policies for DeepLabv3+ as explained earlier, and followed the recommendations of \cite{chen2018encoder} to upscale the logits to input image size for evaluation. Pictures in the Stanford Background dataset have a maximum pixel size of $320 \times 320$ for which we used $321 \times 321$ as crop size as suggested in \cite{chen2018encoder}. 

\textbf{Augmentation, inference, and evaluation.} We augmented the pictures with random flips for both pixel-wise and adversarial training stages. No special inference strategy is applied (no MS, etc.) and the evaluation metric is the mIoU score. In Section~\ref{sec:ablation}, we briefly reflect on and illustrate the impact of applying MS. For early stopping of the discriminator and for mIoU evaluation during adversarial training we used $30\%$ of the validation set of each dataset as the hold-out set.

\textbf{Baseline and performance comparison.} We compare our performance with the non-adversarial DeepLabv3+ as our main baseline and to accentuate on the impact of the proposed conditional adversarial training (LoAd), also with the adversarial approach proposed in \cite{luc2016semantic}. Note that \cite{luc2016semantic} does not employ DeepLabv3+ and we had to reproduce comparable results. The work presented in \cite{hung2018adversarial} could be another option; however, it is focused on \emph{semi-supervised} adversarial segmentation and due to its different scope could not be immediately used for comparison purposes.

\section{Evaluation Results}
\label{sec:res}
\vspace{-0.4cm}
In this section, we evaluate LoAd on the mentioned three datasets. For PASCAL VOC and Cityscpaes, we present performance evaluation results on their \emph{validation} as well as \emph{test} sets, where the latter is assessed by submitting the results to the corresponding online test server. For Stanford Background, the full dataset is available and we only present the results of the test set. Besides the baseline DeepLabv3+, we also compare the quantitative performance with the adversarial approach of \cite{luc2016semantic}. \textbf{Remark:} for viewing the online test results, please note that PASCAL VOC server can be slow sometimes and might require patience or refreshing the page a few times.   
\begin{figure}[!t]
	\centering
	\includegraphics[width=0.6\textwidth]{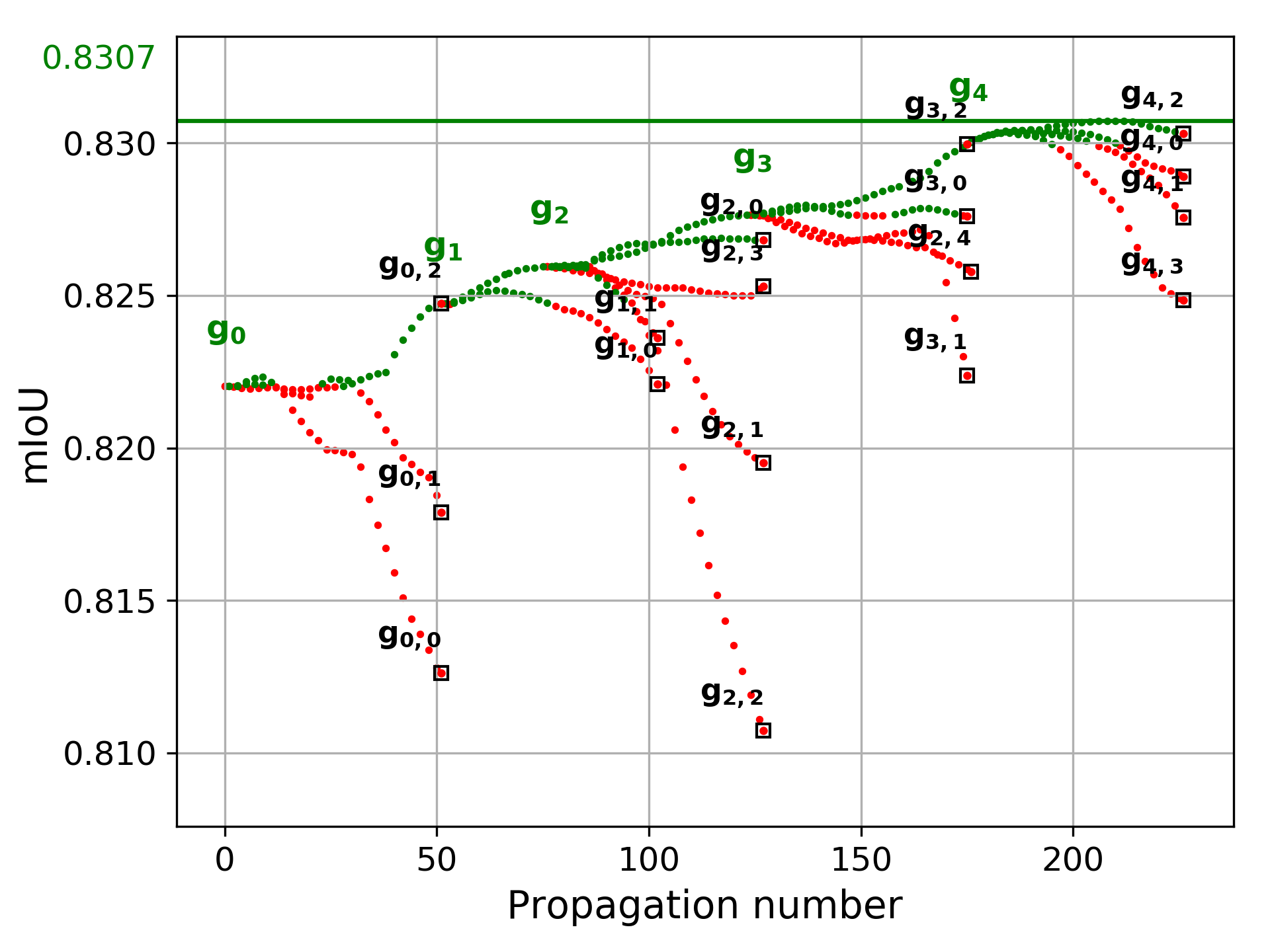}
	\vspace{-0.4cm}
	\caption{Convergence graph of PASCAL VOC 2012.}
	\label{fig:convergence}
	\vspace{-0.2cm}
\end{figure} 

\textbf{Results on PASCAL VOC 2012.} Fig.~\ref{fig:convergence} shows an example of the adversarial training mIoU evaluated on the hold-out set. Note the behavioral similarity to the sketch presented in Fig.\ref{fig:maskaggregation} a). As can be seen, starting from the baseline DeepLabv3+ with mIoU$= 82.2\%$ (denoted by $g_0$), we initially diverge towards $g_{0,0}$ and $g_{0,1}$ in the first two cycles, each time going back to $g_{0}$ and applying map aggregation as discussed in Section~\ref{sec:lookaeahd}. This clearly helped to stabilize the model in the third cycle to ascend from $g_0$ towards $g_{1}$, which is the new peak found. Starting from $g_{1}$ again the network tends to descend towards divergence (ending with $g_{1,0}$) after which applying the map aggregation helped to ascend and find the next peak $g_{2}$. This process continues following the mechanics of LoAd as described in Section~\ref{sec:lookaeahd} alleviating the unstable behavior of the adversarial network helping it to sequentially improve. 

Table~\ref{tb:pascal_val} summarizes the performance comparison between the baseline (DeepLabv3+, also denoted as DLv3+ for brevity), the adversarial approach of \cite{luc2016semantic}, and the proposed boosted model after applying LoAd (DeepLabv3+\,\&\,LoAd) on the full validation set. The results are interesting in that even though the overall mIoU has increased by about $0.9\%$ ($0.7\%$ above \cite{luc2016semantic}) averaged over $21$ classes, we observe considerable improvement in several classes. Note that an \emph{average increase} of $0.9\%$ is not insignificant by itself. In some of the highlighted classes such as ``aeroplane'', ``sofa'', ``diningtable'', and ``boat'' the improvement in IoU ranges from $+2\%$ to $+6\%$ which is quite significant in this regime of performance. Obviously, we degrade in some other classes but only by a fraction of a percent. 

To substantiate our understanding, we evaluated the same three models as in Table~\ref{tb:pascal_val} on the \emph{test} set of PASCAL VOC on its online server. The outcome in Table~\ref{tb:pascal_test} once again corroborates our claim. The overall mIoU improves by $0.7\%$ above both DeepLabv3+ and \cite{luc2016semantic} with even a larger number of classes outperforming the baseline DeepLabv3+, when compared to the validation set. However, less extreme performance jumps can be seen with a maximum of $+3.6\%$ boost for ``sofa'', and $+1\%$ to $+2\%$ improvement for ``boat'', ``dining table'', and so on. Another interesting observation is that on the test set the adversarial approach of \cite{luc2016semantic} outperforms LoAd in a couple of target classes, most notably on ``car''. Check out the results for the baseline DeepLabv3+\footnote{DLv3+: \url{http://host.robots.ox.ac.uk:8080/anonymous/SPNVZZ.html}}, the adversarial approach of \cite{luc2016semantic} \footnote{\cite{luc2016semantic}: \url{http://host.robots.ox.ac.uk:8080/anonymous/C2GEVB.html}}, and the proposed method (DeepLabv3+\,\&\,LoAd) \footnote{DLv3+\&LoAd: \url{http://host.robots.ox.ac.uk:8080/anonymous/CNGCEP.html}} on PASCAL VOC's online server.

\begin{table*}[t]
	\scriptsize
	\caption{Performance comparison on PASCAL VOC 2012 \emph{validation} set.}
	\vspace{-0.0cm}
	\label{tb:pascal_val}
	\centering
	\begin{tabular}{l l l l l l l l l l l l}
		\toprule
		Method     	&backg.	&aero.	&bicycle	&bird	&boat	&bottle 	&bus	&car	&cat	&chair	&cow\\
		\midrule
		DLv3+ 	&95.55			&90.33							&44.23	&89.56			&72.15							&81.11			&{\bf 96.76}	&{\bf 91.37}	&{\bf 94.33}	&51.87	&{\bf 96.08}\\
		\cite{luc2016semantic}	&95.57 &91.94	&{\bf 44.27}			&89.99	&72.83	&{\bf 81.48}	&96.70	&91.21	&94.27	  		&51.92	 		&95.75\\
		DLv3+ \&LoAd &{\bf 95.71}	&\ccell{cellgray}{{\bf 93.31}}	&44.14	&{\bf 90.43}	&\ccell{cellgray}{{\bf74.48}}	&81.37	&96.36	&91.34	&93.73	&\ccell{cellgray}{{\bf53.22}}	&95.59\\	
		\toprule
		\toprule
		contd.	&d.table		&dog	&horse	&m.bike	&person &p.plant	&sheep	&sofa	&train	&tv	&mIoU\\
		\midrule
		DLv3+	&60.14							&92.63	&93.33	&89.23	&90.18	&67.19			&93.75	&61.26	&94.81	&80.27	&82.20\\	
		\cite{luc2016semantic} &60.32 &{\bf 92.65}			&93.31	&{\bf 89.36}			&90.19	&67.35	&{\bf 93.90}			&61.30	&94.79	&80.36	&82.35\\
		DLv3+ \&LoAd &\ccell{cellgray}{{\bf65.86}}	&92.35	&93.34	&89.22	&90.20	&{\bf 67.42}	&93.41	&\ccell{cellgray}{{\bf67.20}}	&{\bf 94.91}	&{\bf 80.81}	&\ccell{cellgray}{{\bf83.08}}\\
		\bottomrule
	\end{tabular}\vspace{-0.1cm}
\end{table*}
\begin{table*}[!t]
	\scriptsize
	\caption{Performance comparison on PASCAL VOC 2012 \emph{test} set (online server).}
	\vspace{-0.0cm}
	\label{tb:pascal_test}
	\centering
	\begin{tabular}{l l l l l l l l l l l l}
		\toprule
		Method     	&backg.	&aero.	&bicycle	&bird	&boat	&bottle 	&bus	&car	&cat	&chair	&cow\\
		\midrule
		DLv3+ 	&93.52	&84.27	&39.70	&86.22	&66.67	&79.68	&{\bf 92.12}	&81.89	&85.84	&42.40	&{\bf82.91}\\
		\cite{luc2016semantic}	&{\bf 93.69}	&{\bf 85.10}	&{\bf 39.81}	&86.39 	&\ccell{cellgray}{\bf 68.04}	&{\bf 80.42}	&92.08	&\ccell{cellgray}{{\bf 84.21}}	&85.34	  		&{\bf 43.33}	 		&80.41\\
		DLv3+ \&LoAd &93.68	&84.91	&39.76	&{\bf 86.48}	&\ccell{cellgray}{{\bf 68.02}}	&80.32	&91.98	&82.70	&\ccell{cellgray}{{\bf 88.18}}	&42.64	&82.74
		
		\\
		\toprule
		\toprule
		contd.	&d.table		&dog	&horse	&m.bike	&person &p.plant	&sheep	&sofa	&train	&tv	&mIoU\\
		\midrule
		DLv3+	&75.20	&84.10	&{\bf 83.64}	&{\bf86.17}	&82.68	&62.54	&{\bf 81.69}	&63.39	&81.60	&{\bf 76.70}	&76.81\\	
		\cite{luc2016semantic} 	&76.96	&81.09	&81.78 &85.72	&82.37	&61.62	&79.07			&\ccell{cellgray}{{\bf 67.24}}	&\ccell{cellgray}{{\bf 83.37}}		&74.95	&76.81\\
		DLv3+ \&LoAd &\ccell{cellgray}{{\bf 77.65}}	&\ccell{cellgray}{{\bf 86.02}}	&83.16	&85.80	&{\bf 82.87}	&{\bf 62.56}	&81.33	&\ccell{cellgray}{{\bf 67.03}}	&\ccell{cellgray}{{\bf 83.18}}	&76.67	&\ccell{cellgray}{{\bf 77.51}}\\
		\bottomrule
	\end{tabular}\vspace{-0.1cm}
\end{table*}
%
%
\begin{table*}[t]
	\scriptsize
	\caption{Performance comparison on Cityscapes \emph{validation} set.}
	\vspace{-0.0cm}
	\label{tb:cityscapes_val}
	\centering
	\begin{tabular}{l l l l l l l l l l l}
		\toprule
		Method     	&road	&sidewalk	&building	&wall	&fence	&pole 	&t.light	&t.sign	&vegetation	&terrain\\
		\midrule
		DLv3+ 		  &97.56	&81.08		&90.20		&38.34							&53.23							&50.27							&60.96		&70.71		&90.91	&59.00\\
		\cite{luc2016semantic} 		  &97.54	&81.02		&90.23		&39.26							&52.77							&50.82							&61.17		&70.94		&90.96	&59.25\\
		DLv3+ \&LoAd  &97.59	&{\bf81.21}	&{\bf90.53}	&\ccell{cellgray}{{\bf47.20}}	&\ccell{cellgray}{{\bf54.38}}	&\ccell{cellgray}{{\bf51.81}}	&{\bf61.54}	&{\bf71.11}	&90.92	&\ccell{cellgray}{{\bf60.66}}\\	
		\toprule
		\toprule
		contd.	&sky		&person	&rider	&car	&truck &bus	&train	&m.cycle	&bicycle	&mIoU\\
		\midrule
		DLv3+			&92.95	&{\bf 76.21}	&52.79		&93.36	&68.83	&76.39		&63.43							&53.70		&72.73		&70.67\\
		\cite{luc2016semantic}			&92.99	&76.13	&52.65		&{\bf93.38}	&{\bf69.85}	&{\bf 77.30}		&62.72							&54.45		&72.76		&70.85\\	
		DLv3+ \&LoAd 	&92.94	&76.13	&{\bf53.27}	&93.19		&67.76		&76.96	&\ccell{cellgray}{{\bf65.00}}	&{\bf54.63}	&{\bf72.85}	&\ccell{cellgray}{{\bf71.57}}\\
		\bottomrule
	\end{tabular}\vspace{-0.1cm}
\end{table*}
\begin{table*}[!t]
	\scriptsize
	\caption{Performance comparison on Cityscapes \emph{test} set (online server).}
	\vspace{-0.0cm}
	\label{tb:cityscapes_test}
	\centering
	\begin{tabular}{l l l l l l l l l l l}
		\toprule
		Method     	&road	&sidewalk	&building	&wall	&fence	&pole 	&t.light	&t.sign	&vegetation	&terrain\\
		\midrule
		DLv3+ 			&97.86	&80.91	&90.14		&43.73							&48.73		&49.14							&62.11		&67.82	&91.69		&66.42\\
		\cite{luc2016semantic} 		  &97.84	&80.79		&90.19		&44.62							&48.22							&49.64							&62.31		&{\bf 67.84}		&91.77	&66.86\\
		DLv3+ \&LoAd 	&97.85	&{\bf 80.99}	&{\bf90.25}	&\ccell{cellgray}{{\bf50.17}}	&{\bf49.46}	&\ccell{cellgray}{{\bf50.24}}	&{\bf62.73}	&67.32		&{\bf91.80}	&\ccell{cellgray}{{\bf68.32}}\\	
		\toprule
		\toprule
		contd.	&sky		&person	&rider	&car	&truck &bus	&train	&m.cycle	&bicycle	&mIoU\\
		\midrule
		DLv3+			&{\bf93.85}	&{\bf79.43}	&59.21	&93.86	&53.41		&64.54							&58.96							&56.79	&{\bf68.55}	&69.85\\	
		\cite{luc2016semantic}			&93.84	&79.36	&59.13		&{\bf 93.87}	&54.28	&66.72		&59.22							&{\bf 57.22}		&68.50		&70.12\\
		DLv3+ \&LoAd 	&93.70		&79.27		&{\bf 59.23}	&93.75		&{\bf54.38}	&\ccell{cellgray}{{\bf67.61}}	&\ccell{cellgray}{{\bf60.24}}	&56.32		&68.36		&\ccell{cellgray}{{\bf70.63}}\\
		\bottomrule
	\end{tabular}\vspace{-0.1cm}
\end{table*}
%
\begin{table*}[!th]
	\footnotesize
	\caption{Performance comparison on Stanford Background \emph{test} set.}
	\vspace{-0.0cm}
	\label{tb:stanford}
	\centering
	\begin{tabular}{l l l l l l l l l l}
		\toprule
		Method     	&sky	&tree	&road	&grass	&water	&building	&mountain	&foreground	&mIoU\\
		\midrule
		DeepLabv3+ 	&89.38			&72.21			&87.28			&77.44			&72.70						&80.03			&48.64						&66.92	&74.33\\
		\cite{luc2016semantic} 	&89.35			&72.54			&87.31			&77.53			&72.78						&80.04			&49.18						&66.69	&74.43\\
		DeepLabv3+ \&LoAd 	&{\bf 89.47}	&{\bf 72.89}	&{\bf 87.71}	&{\bf 77.90}	&\ccell{cellgray}{{\bf73.89}}	&{\bf 80.87}	&\ccell{cellgray}{{\bf 50.11}}	&{\bf 67.54}	&\ccell{cellgray}{{\bf 75.05}}\\
		\bottomrule
	\end{tabular}
\end{table*}
%

\textbf{Results on Cityscapes.} We conducted similar extensive experimentation on Cityscapes as well. Table~\ref{tb:cityscapes_val} summarizes the performance comparison between the baseline DeepLabv3+ (abbreviated as DLv3+), the adversarial approach of \cite{luc2016semantic}, and the proposed boosted model DeepLabv3+\&LoAd. Here again we achieve an overall mIoU improvement of $0.9\%$ ($0.7\%$ above \cite{luc2016semantic}) averaged over $19$ classes with considerable performance boost (up to $+7\%$) in ``wall'', ``fence'', ``pole'', ``train'' and ``terrain''. To further consolidate our understanding, we evaluated the same three models as in Table~\ref{tb:cityscapes_val} on the \emph{test} set of Cityscapes on its online server and the results are summarized in Table~\ref{tb:cityscapes_test}. We achieve an overall performance boost of about $0.8\%$ above the baseline DeepLabv3+ ($0.5\%$ above \cite{luc2016semantic}) and a considerable improvement (up to $+5\%$) for ``wall'', ``terrain'', ``train'', and so on, that once again confirm the impact of the proposed method on this commonly used dataset. Here, we consistently outperform the adversarial approach of \cite{luc2016semantic} on the aforementioned classes. Check out the results for the baseline DeepLabv3+\footnote{DLv3+: \url{https://tinyurl.com/ctyscps-dlv3plus}}, the adversarial approach of \cite{luc2016semantic} \footnote{\cite{luc2016semantic}: \url{https://tinyurl.com/ctyscps-dlv3plus-luc-2}}, and the proposed method \footnote{DLv3+\&LoAd: \url{https://tinyurl.com/ctyscps-dlv3plus-load-2}} on Cityscapes' online server.

\textbf{Results on Stanford Background.} Following \cite{luc2016semantic}, we conducted yet another set of experimentation on the Stanford Background dataset. The results are summarized in Table~\ref{tb:stanford}. As can be seen, applying LoAd on DeepLabv3+ boosts the overall mIoU by $0.7\%$ ($0.6\%$ beyond \cite{luc2016semantic}) averaged over $8$ classes, which is slightly less than PASCAL VOC 2012 and Cityscapes. However, here the boosted model consistently outperforms the baseline DeepLabv3+ and the adversarial approach of \cite{luc2016semantic} on all classes without any exception with considerable improvement on ``mountain'' and ``water''.


\begin{figure*}[!th]
	\centering
	\includegraphics[width=.87\textwidth]{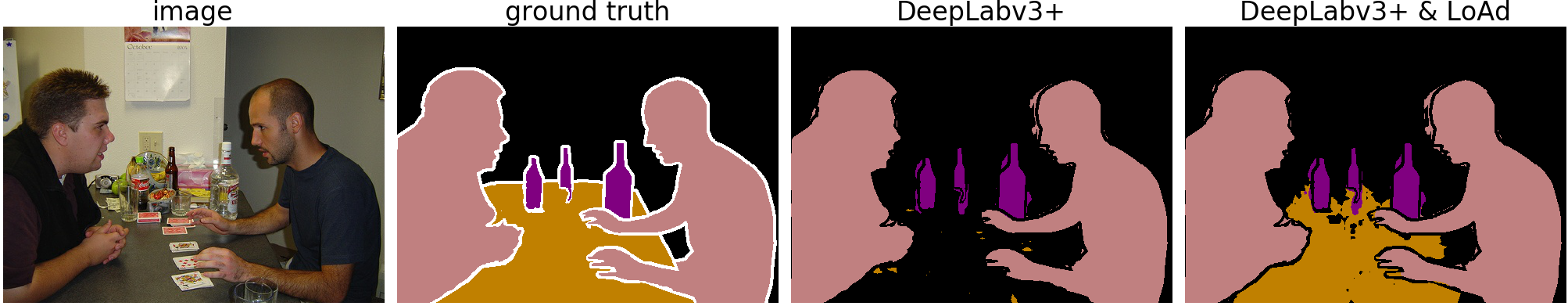}
	\\[\smallskipamount]
	\vspace{-0.1cm}
	\includegraphics[width=.87\textwidth]{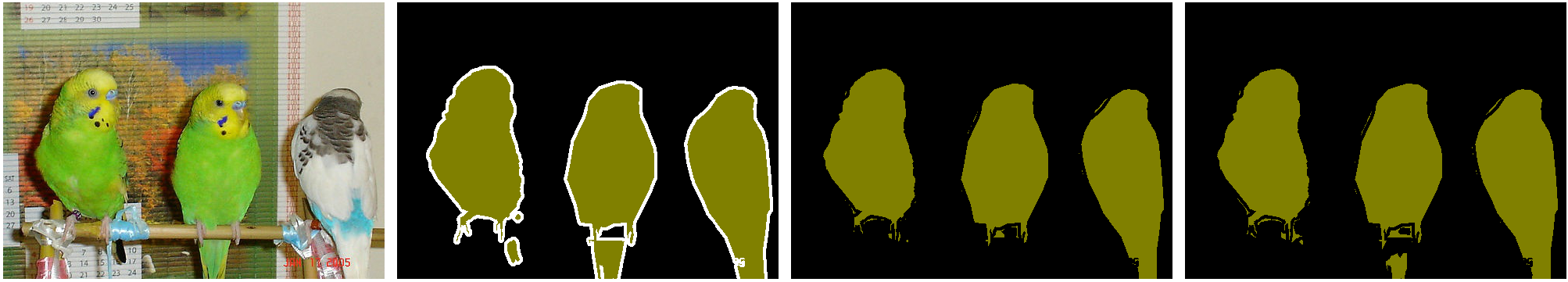}
	\\[\smallskipamount]
	\vspace{-0.1cm}
	\includegraphics[width=.87\textwidth]{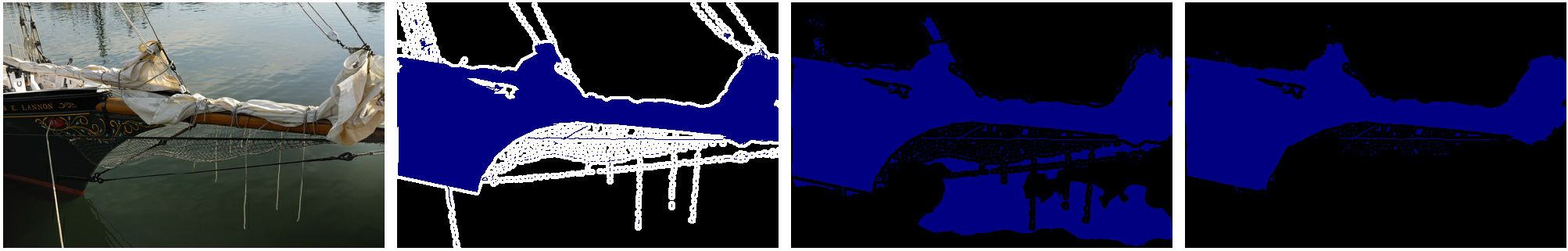}
	\\[\smallskipamount]
	\vspace{-0.1cm}
	\includegraphics[width=.87\textwidth]{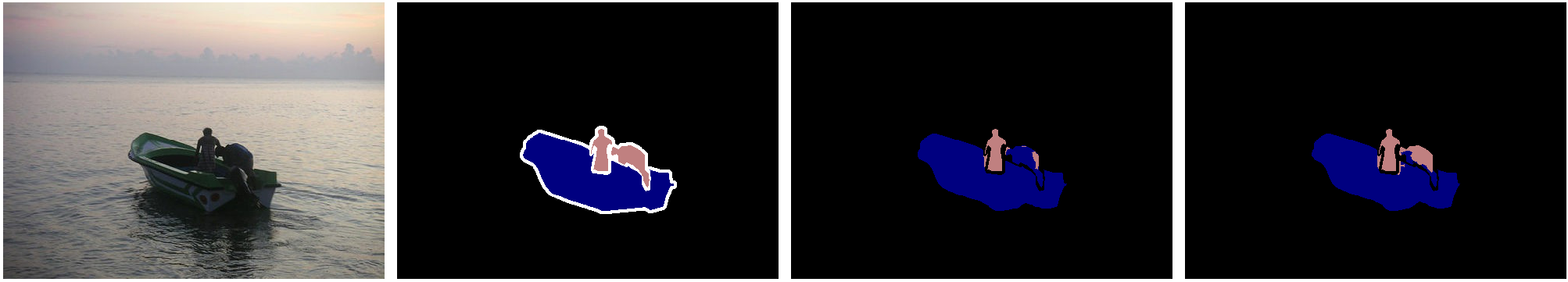}
	\\[\smallskipamount]
	\vspace{-0.1cm}
	\hspace{-0.1cm}
	\includegraphics[width=.87\textwidth]{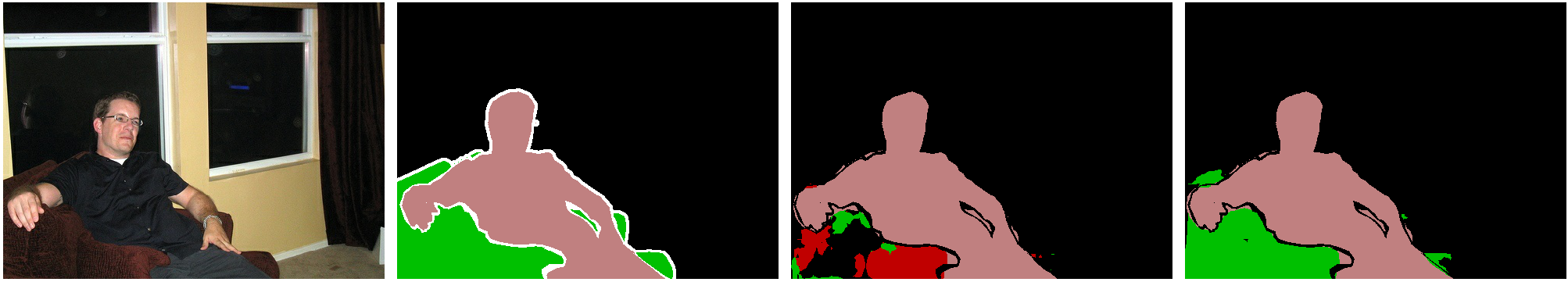}
	\\[\smallskipamount]
	\vspace{-0.07cm}
	\includegraphics[width=.87\textwidth]{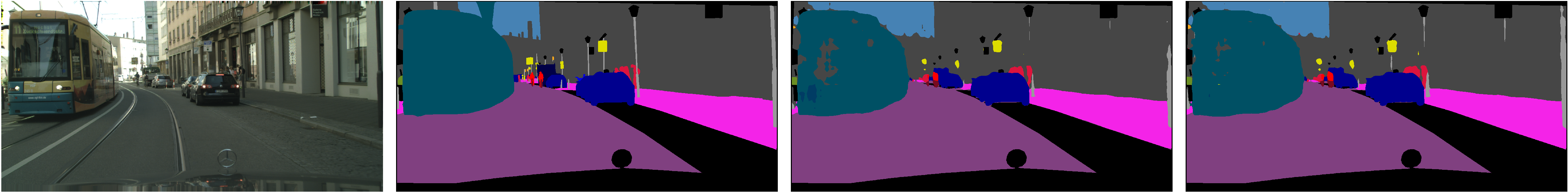}
	\\[\smallskipamount]
	\vspace{-0.07cm}
	\includegraphics[width=.87\textwidth]{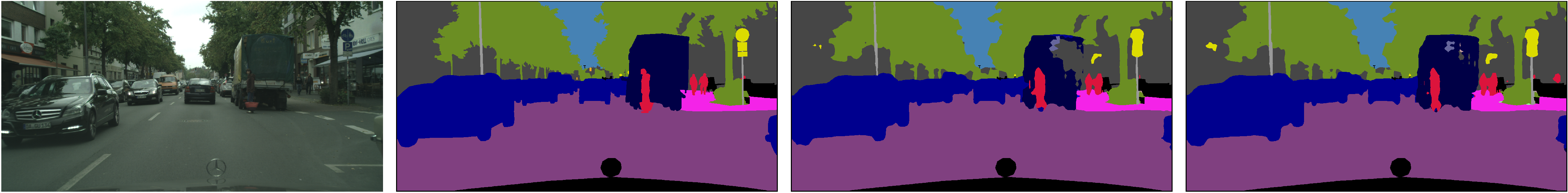}
	\\[\smallskipamount]
	\vspace{-0.2cm}

	\caption{Selected qualitative results on PASCAL VOC 2012 and Cityscapes. These examples show how totally missing items and class swap/confusion are resolved or considerably improved. Ignored pixels and classes (per standard) unused for evaluation in black. Best view in color with $300\%$ zoom.}
	\label{fig:comparison_1}
	\vspace{-0.3cm}
\end{figure*}

\textbf{Qualitative results.} A selected set of qualitative results are illustrated in the next four figures. In Fig~\ref{fig:comparison_1}, on the top row, the whole dining table is missed by the baseline model and LoAd manages to boost DeepLabv3+ to fully recover that. ``Bird'' tails are interesting example of how discontinuity is less of an impediment for the proposed method. On the third row, once again the impact is significant, a misconception of a ``boat'' in the middle of the water is resolved. Th next three rows show interesting signs of avoiding class swap/confusion. Being a rather difficult case to distinguish even for human observer, LoAd helps DeeLabv3+ to resolve mistaking the second ``person'' on the boat with boat engine on the fourth row. The same can be argued about the next two rows where ``sofa'' vs. ``chair'' (in red) and ``train'' vs. ``building'' (in dark gray) confusions are resolved, respectively. The last row, towards the right side of the image, demonstrates a much more consistent segmentation of ``truck'' as compared to the baseline DeepLabv3+.
\begin{figure*}[!th]
	\centering
	\includegraphics[width=.87\textwidth]{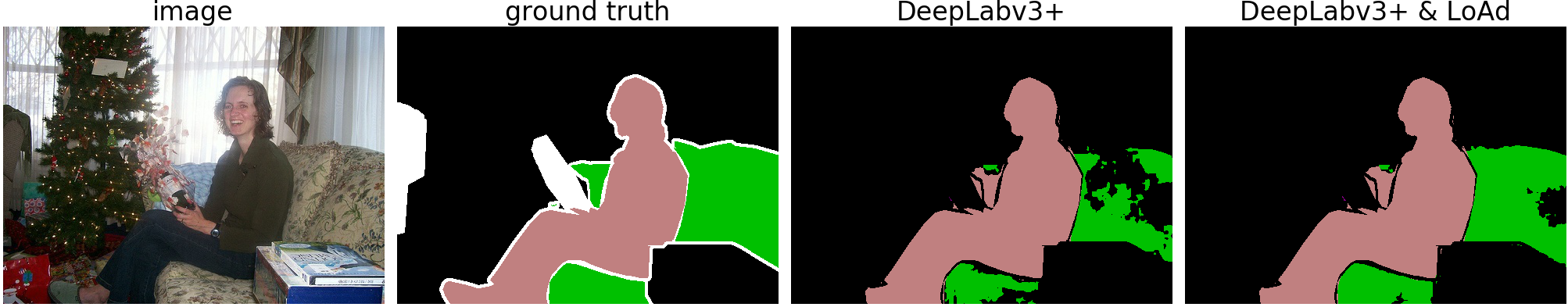}
	\\[\smallskipamount]
	\vspace{-0.1cm}
	\includegraphics[width=.87\textwidth]{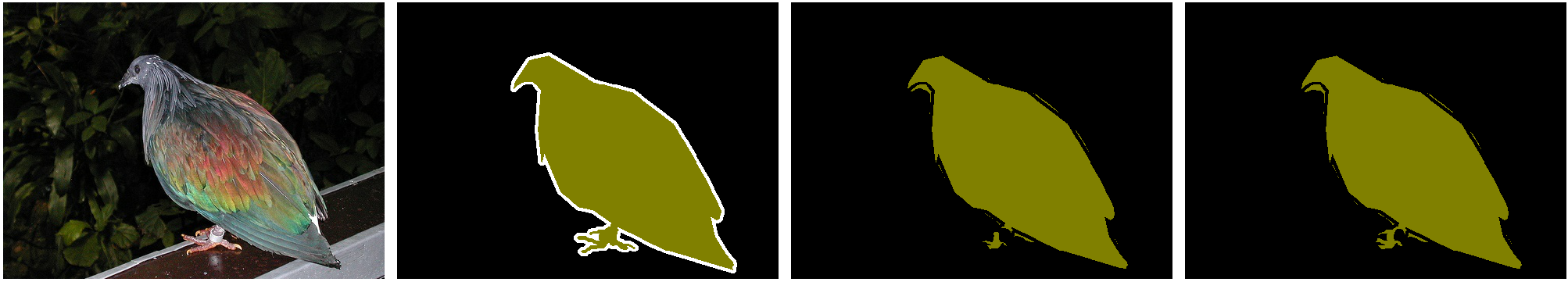}
	\\[\smallskipamount]
	\vspace{-0.1cm}
	\includegraphics[width=.87\textwidth]{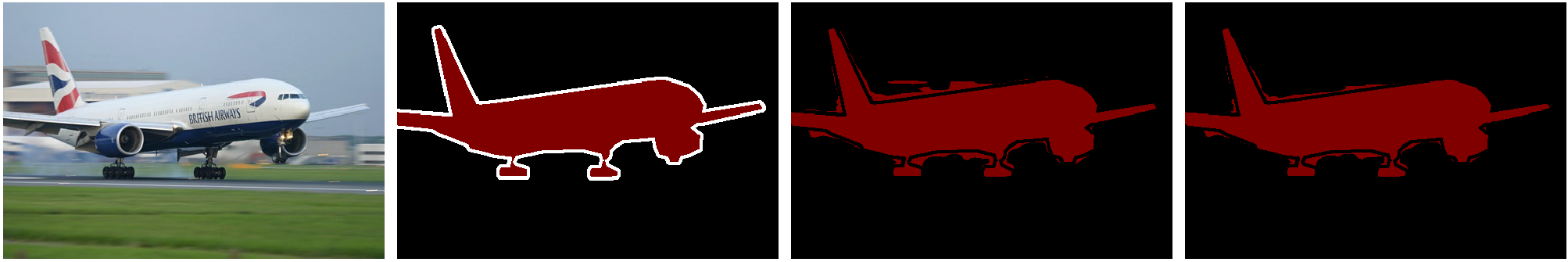}
	\\[\smallskipamount]
	\vspace{-0.1cm}
	\includegraphics[width=.87\textwidth]{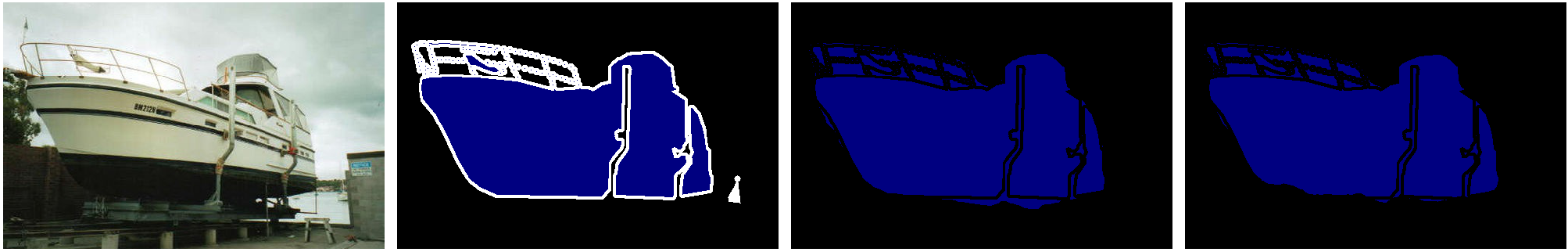}
	\\[\smallskipamount]
	\vspace{-0.1cm}
	\includegraphics[width=.87\textwidth]{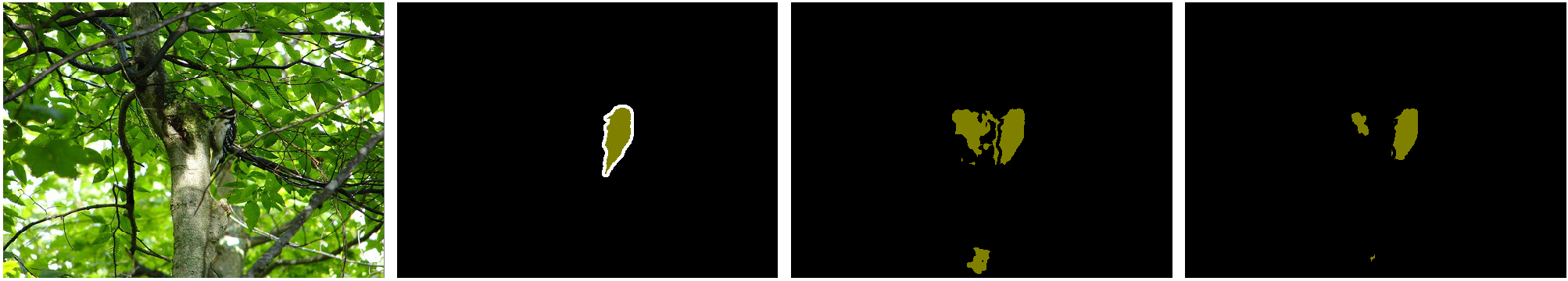}
	\\[\smallskipamount]
	\vspace{-0.05cm}
	\includegraphics[width=.87\textwidth]{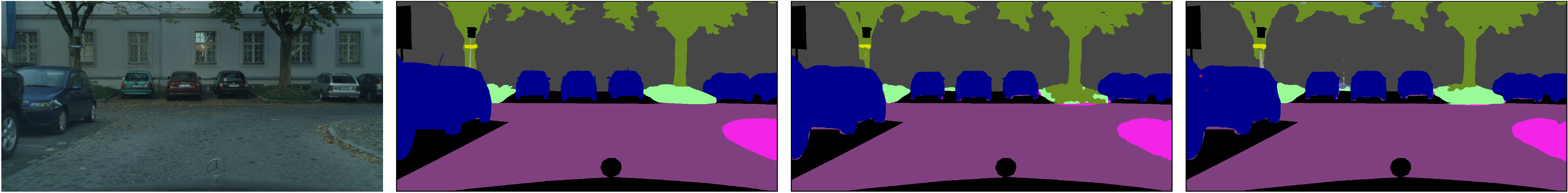}
	\\[\smallskipamount]
	\vspace{-0.05cm}
	\includegraphics[width=.87\textwidth]{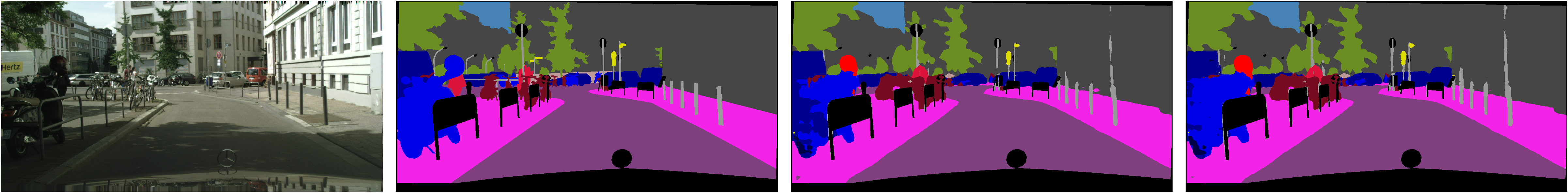}
	\\[\smallskipamount]
	\vspace{-0.2cm}
	\caption{Selected qualitative results on PASCAL VOC 2012 and Cityscapes. These examples demonstrate how inconsistent body/edges and class swap/confusion are either totally resolved or considerably improved. Ignored pixels and classes (per standard) unused for evaluation in black. Best view in color with $300\%$ zoom.}
	\label{fig:comparison_2}
\end{figure*}

\newpage
\begin{figure*}[!th]
	\centering
	\includegraphics[width=0.72\textwidth]{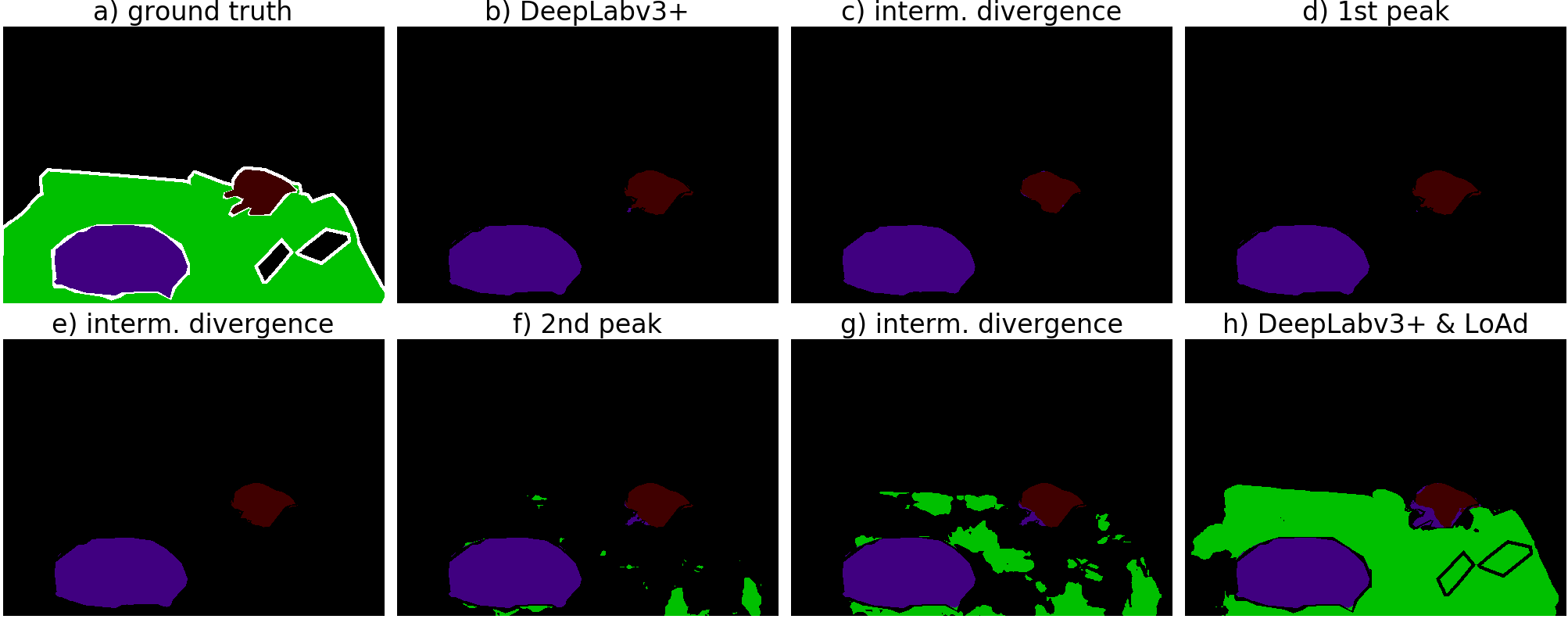}
	\\[\smallskipamount]
	\vspace{+0.1cm}
	\includegraphics[width=0.72\textwidth]{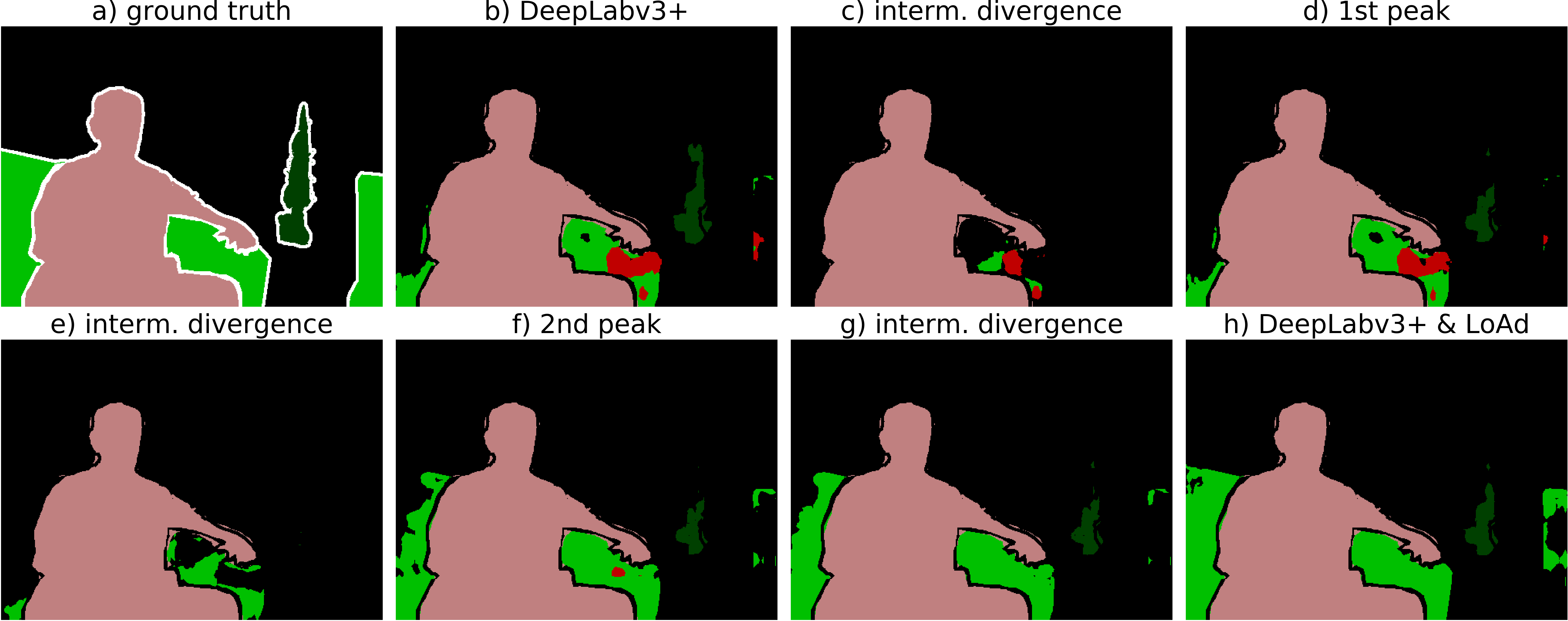}
	\\[\smallskipamount]
	\vspace{+0.1cm}
	\includegraphics[width=0.72\textwidth]{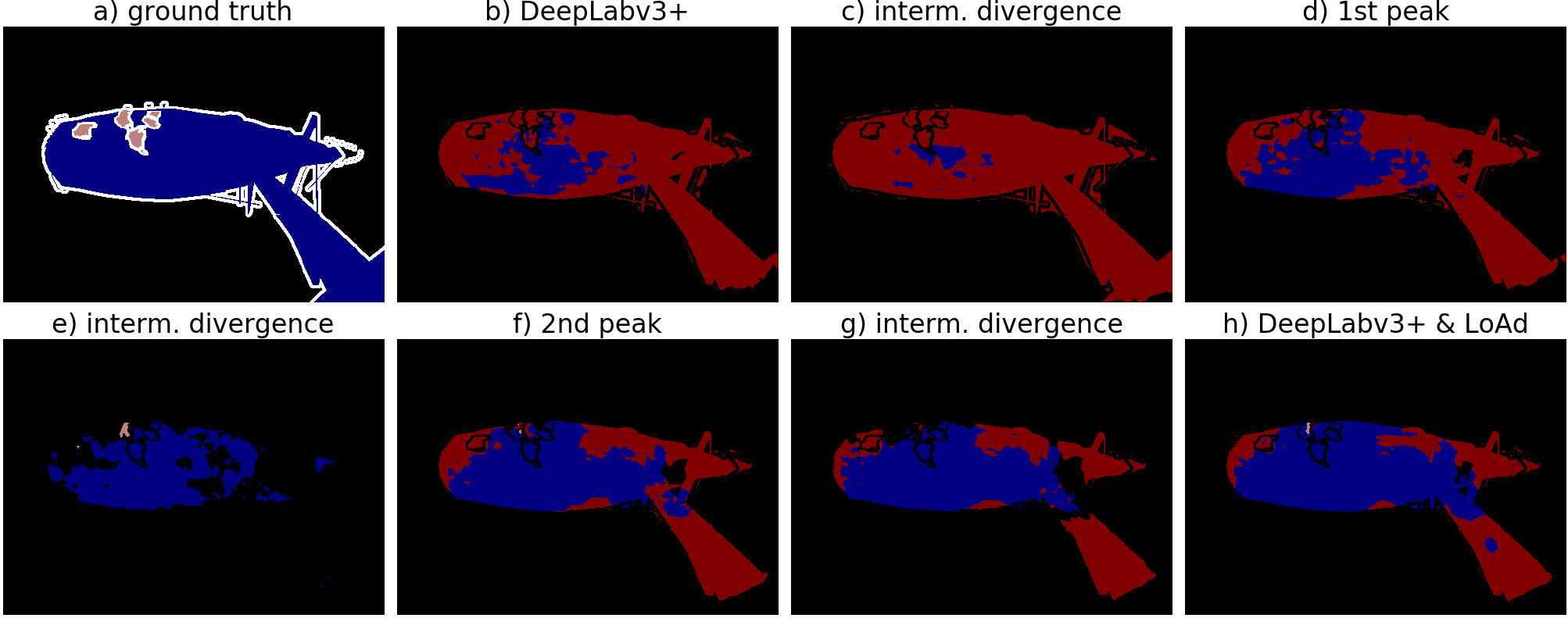}
	\\[\smallskipamount]
	\vspace{+0.1cm}
	\includegraphics[width=0.72\textwidth]{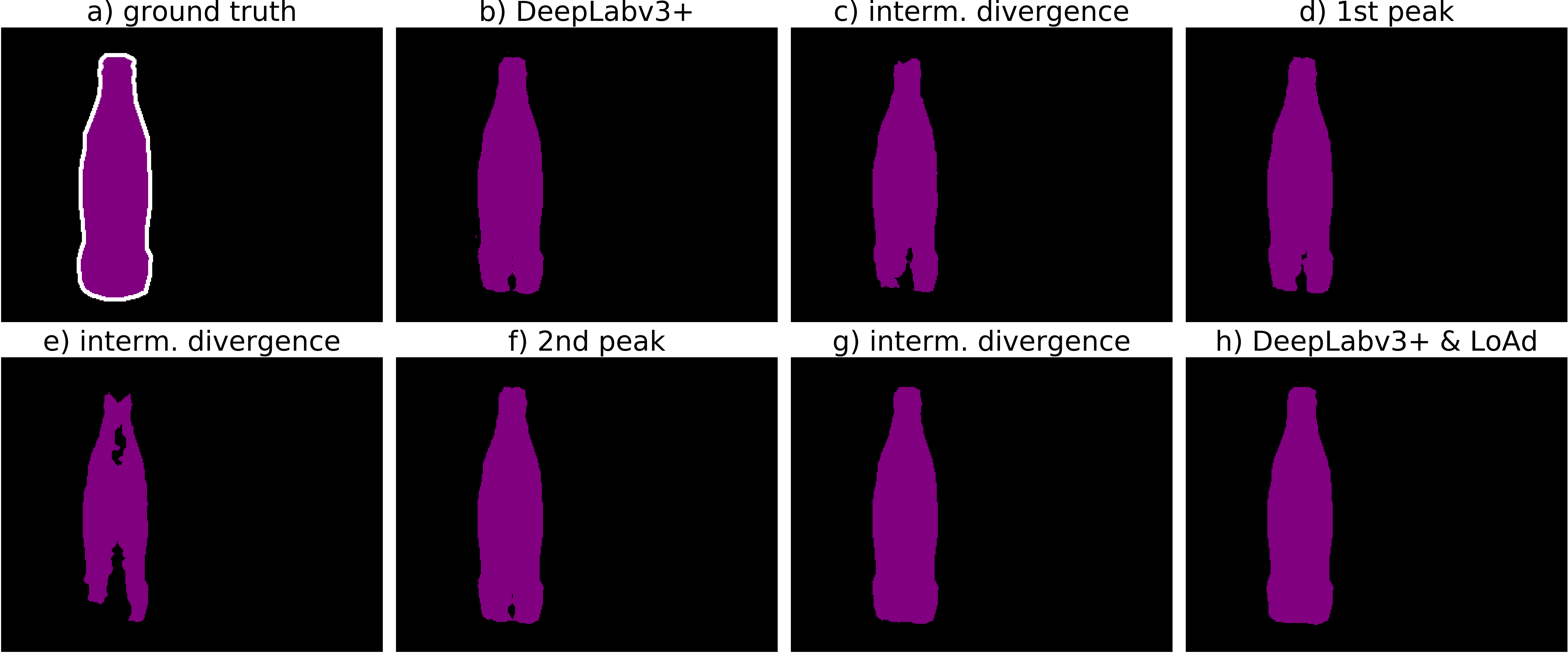}
	\vspace{-0.2cm}
	\caption{This illustrates how LoAd benefits from label map aggregation throughout its convergence process to boost the performance of the baseline DeepLabv3+. Every consecutive two rows correspond to a single input image. Best view in color with $300\%$ zoom.}\label{fig:recovery_supp_1}
\end{figure*}

\newpage
\begin{figure*}[!th]
	\centering
	\includegraphics[width=.8\textwidth]{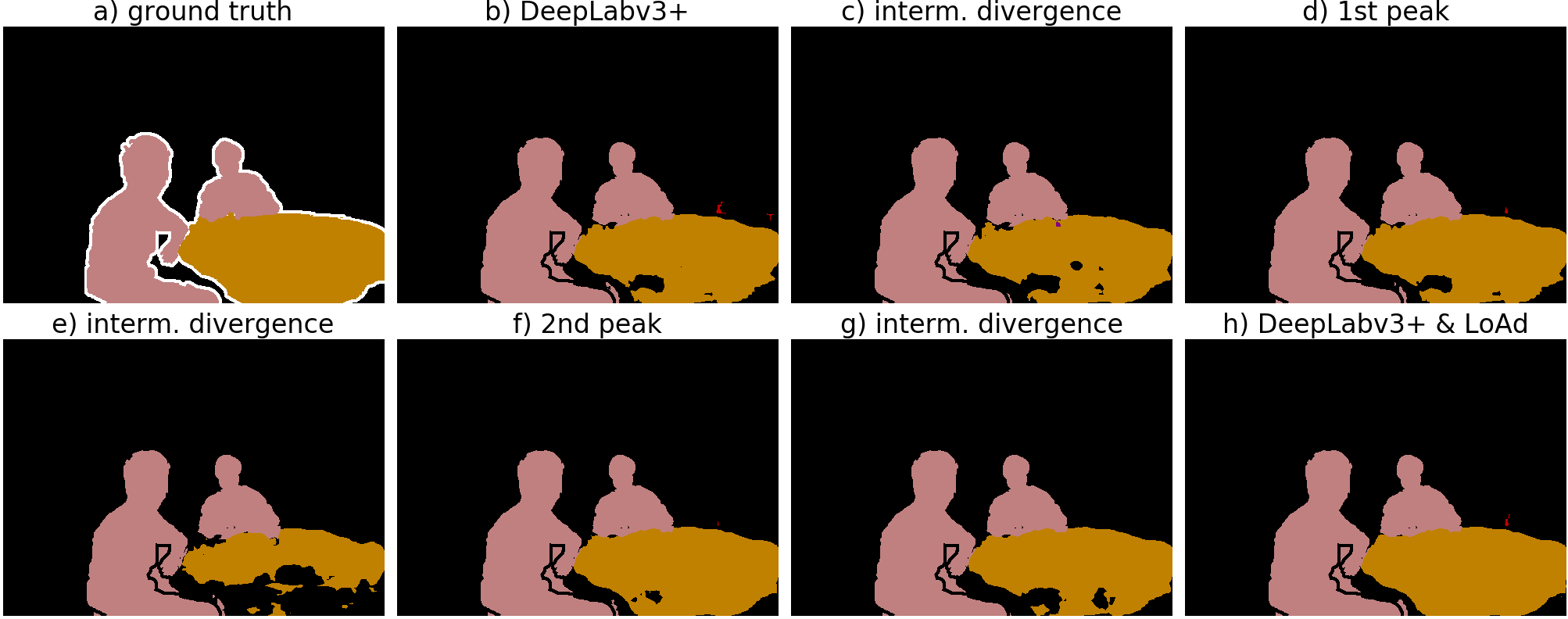}
	\\[\smallskipamount]
	\vspace{+0.1cm}
	\includegraphics[width=.8\textwidth]{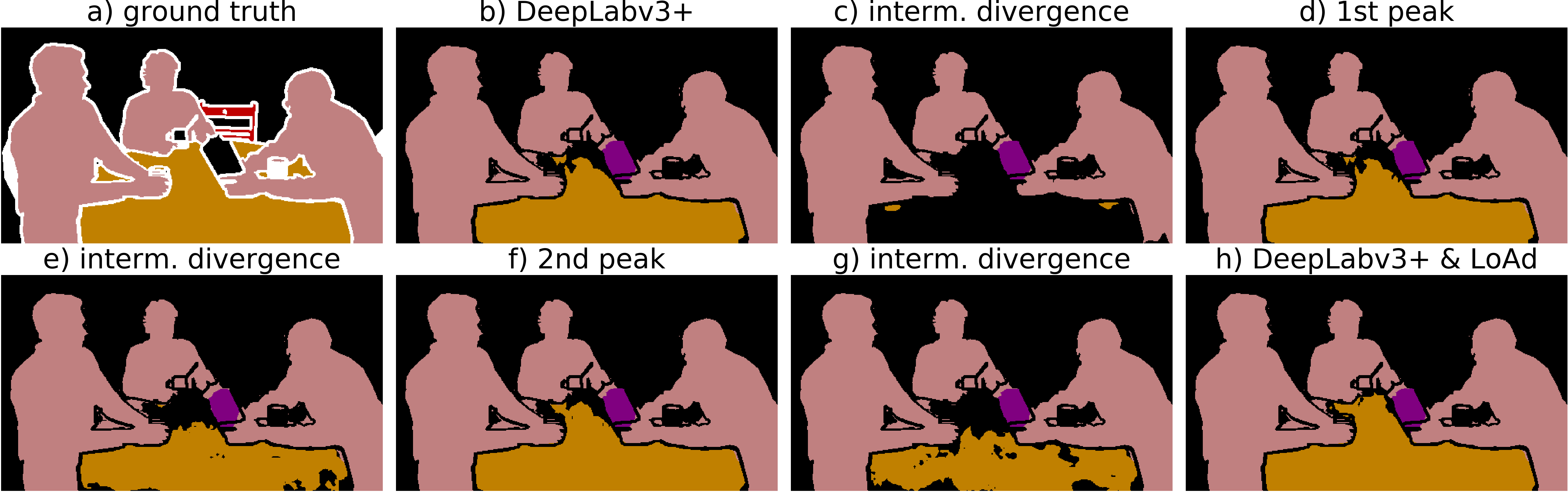}
	\\[\smallskipamount]
	\vspace{+0.1cm}
	\includegraphics[width=.8\textwidth]{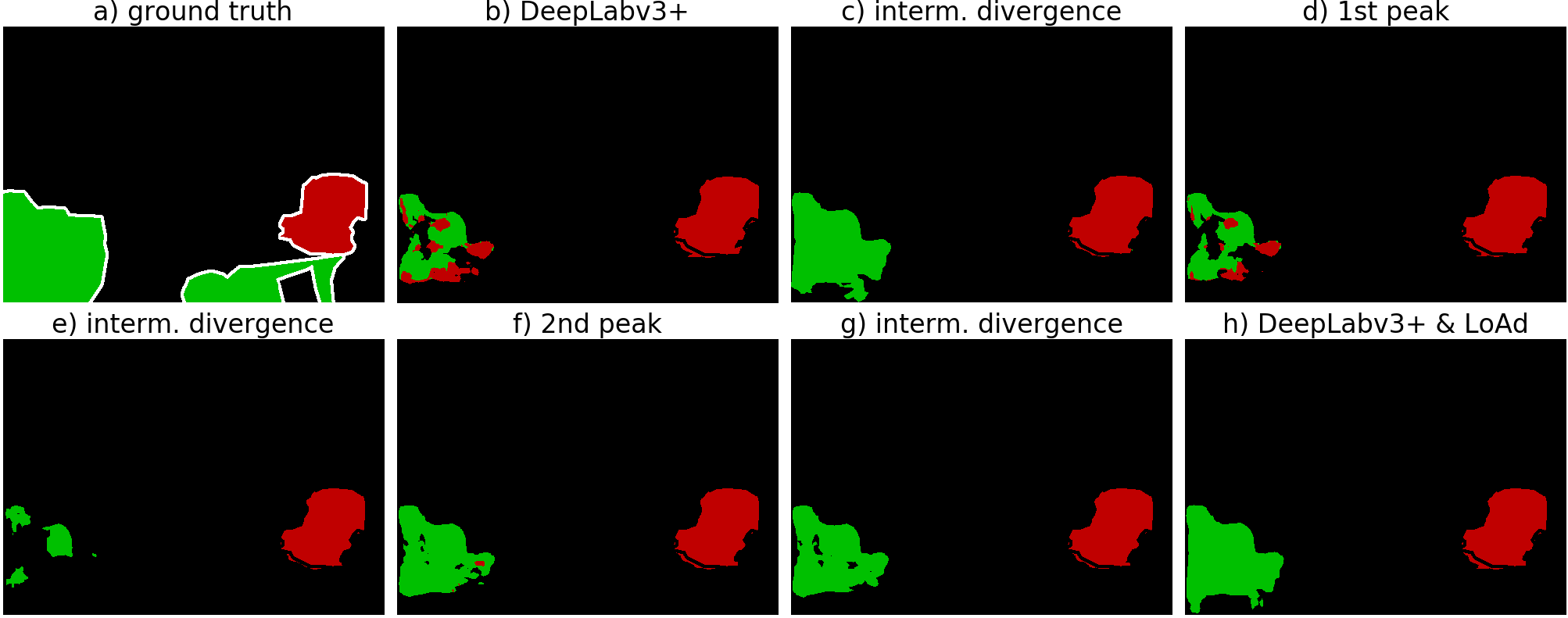}
	\\[\smallskipamount]
	\vspace{+0.1cm}
	\includegraphics[width=.8\textwidth]{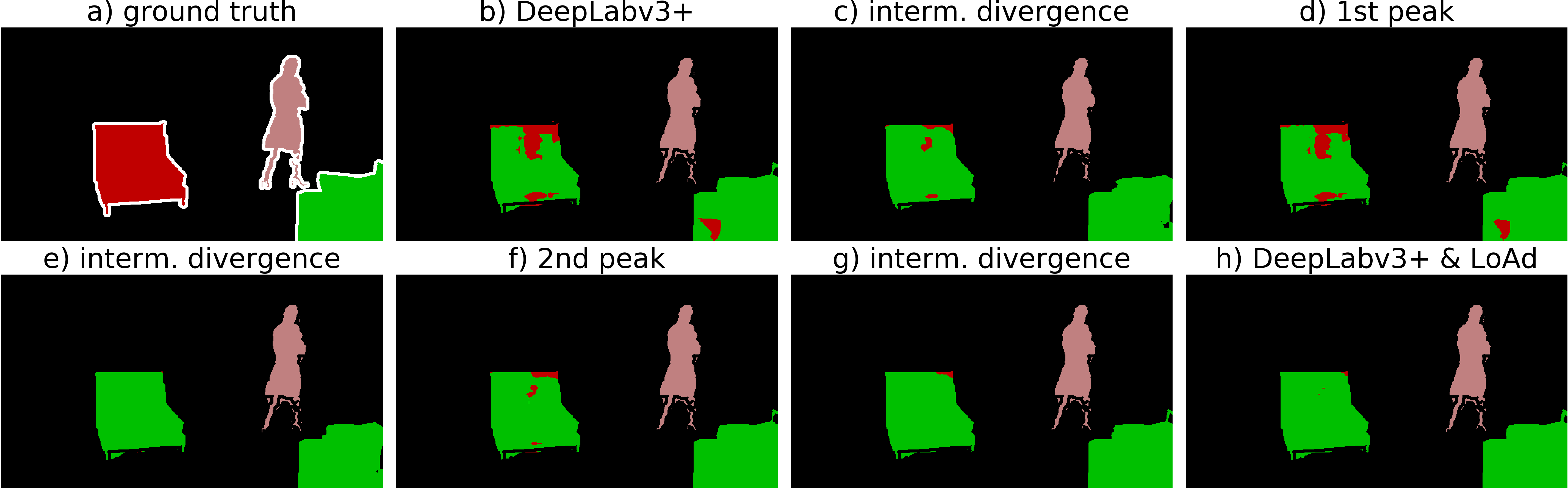}
	\\[\smallskipamount]
	\vspace{-0.2cm}
	\caption{This illustrates how LoAd benefits from label map aggregation throughout its convergence process to boost the performance of the baseline DeepLabv3+. Every consecutive two rows correspond to a single input image. Best view in color with $300\%$ zoom.}\label{fig:recovery_supp_2}
\end{figure*} 
Another set of qualitative results from both PASCAL VOC and Cityscapes datasets is illustrated in Fig~\ref{fig:comparison_2}.On the top row, ``sofa'' is segmented with better consistency and continuity. The second row illustrates the continuity in segmenting ``bird'' foot. The third and and fourth rows depict cleaner edges at the bottom and around ``aeroplane'' and ``boat''. The fifth row shows how LoAd outperforms in detecting the ``bird'' in the cluttered scene, and manages to (a good extent) resolve the confusion of tree trunk with another bird in the scene. The sixth row, in the middle of the image towards right, depicts improvement in distinguishing ``tree'' and ``terrian'' (in two shades of green). The last row, on the very left side of the image, shows considerably better performance is segmenting ``motorcycle''  and avoiding confusion with ``car'' (in two shades of blue). 

To further consolidate the qualitative improvement of DeepLabv3+\,\&\,LoAd compared to DeepLabv3+, we provide eight examples in Figs.~\ref{fig:recovery_supp_1} and \ref{fig:recovery_supp_2} where the gradual impact of LoAd with its embedded label map aggregation on the performance of the baseline throughout its cycles (ending up with divergence or leading to finding new peaks) is at the center of attention. As can be seen, starting from the baseline DeepLabv3+, LoAd gradually learns from the intermediate degradation in its downward divergent stages at c), e) and g) and resolves confusion. The first example (top two rows) in Fig.~\ref{fig:recovery_supp_1} is a prime example of how a complete instance of ``sofa'' is missed by the baseline DeepLabv3+ (b), whereas LoAd manages to completely recover it (h). The second example (next two rows) reiterate the same message, offering much better consistency compared to the baseline. The third example (third two rows) poses a challenging scenario where a ``boat'' and an ``aeroplane'' are confused due to perspective and texture complexity. LoAd (h) shows a considerable improvement in resolving this confusion. Last example in this figure depicts how the missing bottom part of the bottle is recovered owing to label map aggregation. The same story continues in Fig.~\ref{fig:recovery_supp_2}. Top two examples highlight how LoAd helps the baseline to infill the gaps and missing pieces in segmenting a ``dinning table''. Last two examples (last four rows) in the figure demonstrate how class confusion (``sofa'' vs ``chair'') is totally resolved when compared to the baseline. 

\clearpage
\section{Ablation Study}
\label{sec:ablation}
\vspace{-0.4cm}
The first angle to investigate is the impact of changing the core components of the adversarial network. Our initial experimentation with VGG$16$ as well as custom designed CNNs as discriminator did not lead to any notable performance boost or degradation. So, here we focus on the impact of modifying the generator (segmentor), currently being DeepLabv3+. Table~\ref{tb:bakcbone} summarizes the results of our experimentation with a different backbone (MobileNetv2) for the segmentor in comparison with the main backbone used in our previous experimentation (Xception-$65$). The results reported in this table correspond to the best performing models (baseline and proposed) on PASCAL VOC 2012, for the sake of a fair comparison. Obviously, a weaker backbone (MobileNetv2) results in performance drop for both baseline and boosted model with LoAd on \emph{validation} and \emph{test} datasets. Nevertheless, what remains consistent is the improvement of DeepLabv3+\,\&\,LoAd over the baseline, even though the improvement is a bit less pronounced for both \emph{validation} and \emph{test} datasets. 


Another angle to investigate is the impact of hyperparamter change on the performance of DeepLabv3+\,\&\,LoAd. We perturb all the important design hyperparameters of LoAd described in Section~\ref{sec:lookaeahd} and in Algorithm~\ref{alg:lookahead_simplified}. The results summarized in Table~\ref{tb:hyper} are extracted with our main choice of backbone (Xception-$65$) and each averaged over $3$ experiments. As can be seen, increasing the maximum set size of the buffer results in performance degradation of DeepLabv3+\&\,LoAd in comparison with the best setting with $B_\text{max} = 3$. This could potentially be due to the fact that we only generate and aggregate ``fake'' label map datasets, keeping only one ground truth label map set, which can be a source of imbalance if too many of these sets are aggregated. This is the exact reason why we have introduced this cap, $B_\text{max}$. A possible solution to remedy this issue, besides weighting the discriminator loss as discussed in Section~\ref{sec:exp}, could be to generate fake label maps only for a subset of training image set, e.g., images that have been seen by the corresponding model (peak or ending) up to that stage of training.
\begin{table}[t]
	\footnotesize
	\caption{Impact of generator backbone change for \emph{val} and \emph{test} results of PASCAL VOC 2012.}
	\vspace{-0.0cm}
	\label{tb:bakcbone}
	\centering
	\begin{tabular}{l l l l}
		\toprule
		Method     	&Gen. backbone &mIoU (\emph{val}) &mIoU (\emph{test})	\\
		\midrule
		DLv3+ 	&Xception-65				&82.20			&76.81\\
		DLv3+\,\&\,LoAd 	&Xception-65	&{\bf83.08}  	&{\bf77.51}\\
		\midrule
		DLv3+ 	&MobileNetv2				&75.32			&71.45\\
		DLv3+\,\&\,LoAd 	&MobileNetv2	&{\bf75.70}   	&{\bf71.74}\\
		\bottomrule
	\end{tabular}\vspace{-0.3cm}
\end{table}
\begin{table}[t]
	\footnotesize
	\caption{Influence of hyperparameter change on LoAd (Xception-$65$) for \emph{val} results of PASCAL VOC 2012.}
	\vspace{-0.0cm}
	\label{tb:hyper}
	\centering
	\begin{tabular}{l l l l l l l}
		\toprule
		Method       &$\beta_l (\%)$ &$\beta_u (\%)$ &$\Gamma$ &$B_\text{max}$	&mIoU (\emph{val})\\
		\midrule
		DLv3+ 			&- 	 	&-      &- 	 	&-	 	&82.20\\
		DLv3+\,\&\,LoAd 	&5 	 	&0.1	&50		&3  	&{\bf 82.88}\\
		DLv3+\,\&\,LoAd 	&5 	 	&0.1 	&50		&4		&82.46\\
		DLv3+\,\&\,LoAd 	&5 	 	&0.5 	&50		&3		&82.42\\
		DLv3+\,\&\,LoAd 	&10 	&0.1 	&50		&3		& 82.80\\
		DLv3+\,\&\,LoAd 	&20 	&0.1 	&50		&3		&82.43\\
		DLv3+\,\&\,LoAd 	&20 	&0.1 	&80		&3		&82.65\\
		\bottomrule
	\end{tabular}\vspace{-0.2cm}
\end{table}

Next experiments are concerned with perturbing $\beta_l$ and $\beta_u$. As can be seen in Table~\ref{tb:hyper}, lowering $\beta_l$ to $10\%$ (in mIoU) does not seem to have a major impact on the overall mIoU ($=82.80\%$), whereas increasing $\beta_u$ to $0.5\%$ downgrades the performance (mIoU $= 82.42\%$), suggesting a conservative upper-bound for finding a peak per cycle. To further investigate this, we conducted two more experiments. The results indicate that decreasing $\beta_l$ further down to $20\%$ seems to gradually degrade the performance (mIoU $= 82.43\%$), even though increasing the divergence patience parameter $\Gamma$ from $50$ to $80$ iterations helps to avoid this to some extent (mIoU $= 82.65\%$). Overall, it seems aggregating heavily degraded label maps (corresponding to lower $\beta_l$'s) does not help the performance as this is implicitly related to how we constrain the solution space of the optimization problem per cycle of LoAd.

Finally, we would like to shed some light on what would happen in terms of performance if we had also employed expensive time-consuming post-processing steps at inference time. Table~\ref{tb:multi_scale} summarizes the impact of applying multi-scale and flip (denoted by MS\&F) \emph{only for evaluation} (the last column) against already reported results without employing MS\&F (the third column). We observe about $1\%$ to $3\%$ performance boost in almost all models pushing their performance towards state-of-the-art results in the literature. However, this performance boost comes at the cost of scales of magnitude (roughly $100$ times) larger inference time rendering the models out of real-time inference response. In summary, pushing the performance at any cost is not our focus; keeping the inference time unchanged (in practice less than $1$ sec) and improving the performance is what we offer in this work. 
\begin{table}[t]
	\footnotesize
	\caption{Impact of multi-scale and flip (MS\&F) inference strategy on validation data (unless otherwise stated) for each dataset. Xception-65 is used as the generator backbone for PASCAL VOC 2012 and Stanford Background, and MobileNetv2 is used for Cityscapes.}
	\vspace{0.1cm}
	\label{tb:multi_scale}
	\centering
	\begin{tabular}{l l l l}
		\toprule
		Method     	&Dataset (\emph{val}) &mIoU  &mIoU (MS\&F)	\\
		\midrule
		DLv3+ 	&VOC 2012				&82.20			&83.58\\
		\cite{luc2016semantic} 	&VOC 2012	&82.35  	&83.66\\
		DLv3+\,\&\,LoAd 	&VOC 2012	&{\bf83.08}  	&{\bf84.57}\\
		\midrule
		DLv3+  	&Cityscapes				&70.67			&73.88\\
		\cite{luc2016semantic} 	&Cityscapes	&70.85   	&73.96\\
		DLv3+\,\&\,LoAd 	&Cityscapes	&{\bf71.57}   	&{\bf74.61}\\
		\midrule
		DLv3+ 	&Stanford (\emph{test})				&74.33			&76.33\\
		\cite{luc2016semantic} 	&Stanford	(\emph{test})&74.43   	&76.42\\
		DLabv3+\,\&\,LoAd 	&Stanford (\emph{test})	&{\bf75.05}   	&{\bf76.78}\\
		\bottomrule
	\end{tabular}\vspace{-0.3cm}
\end{table}
%
\section{Concluding Remarks}
\label{sec:conc}
\vspace{-0.4cm}
\textbf{Summary and impact.} We proposed a novel lookahead adversarial learning (LoAd) approach with an embedded label map aggregation idea for adversarial semantic segmentation with state-of-the-art models. For experimental evaluation, we picked DeepLabv3+ (also abbreviated as DLv3+) as segmentor/generator in the adversarial setting. We further elaborated in Sections~\ref{sec:intro} and \ref{sec:exp} that among a few available choices for DeepLabv3+ architectures and corresponding training/inference strategies, we deliberately picked the ones which can be trained on \emph{commodity GPU nodes} (because not everyone has access to high-end TPUs) and \emph{run fast at inference}; roughly speaking, less than a second on average per sample input image. Therefore, we did not opt for models requiring multi-scaling strategies or small output strides leading to several minutes or in some cases few tens of minutes inference time per image. Running fast at inference time for field applications is a \emph{core motive} of this work. We demonstrated that the proposed idea can improve the performance of the baseline DeepLabv3+ by about $0.9\%$ in mIoU sense on two of the most modern and commonly-used datasets (PASCAL VOC 12, Cityscapes) while boosting the performance in certain classes up to $7\%$ without introducing any extra delay at inference. In other words, the proposed models run practically as fast as the baseline on top of which they are applied. The qualitative results show that our approach is helping the baseline segmentation model to resolve class confusion to a good extent as well as to produce label maps which are more consistent in terms of continuity and structure. We think the application domain of LoAd could possibly be larger than semantic segmentation. This is an interesting avenue to explore. 

\textbf{Complexity.} LoAd adds architectural complexity by adding a (simple) discriminator network but requires no changes to the base segmentation model, as such in theory, it can be applied on top of any semantic segmentation model (not only DeepLabv3+). On the flip side, we introduce complexity at training both in time and space slowing it down in favor of performance. Roughly speaking, in worst case scenario of large size images containing $N$ pixels, for an evaluation (hold-out) subset of size $V_e$ samples, we impose an extra $\mathcal{O}(V_e N)$ time complexity per propagation for adversarial training. In practice, $V_e \ll N$. Also, with the label map aggregation module of LoAd we need to train the discriminator on $B_\text{max}\,M$ images stacked in the buffer (space complexity) instead of only $M_b$ images in a batch with $M_b < M$. On the other hand, we train the discriminator once per LoAd cycle, i.e., orders of magnitude less often than typical GDA based adversarial training. In terms of time complexity, and for the sake of comparison, on a P100 Tesla GPU for PASCAL VOC 2012, DeepLabv3+ takes 20:01:41 (hour/min/sec) and DLv3+\,\&\,LoAd takes 53:48:30 (hour/min/sec) to reach best performance. LoAd's elapsed time is about $2.5$ times larger, even though looking at typical training times in literature (sometimes taking weeks), only $1.5$ days extra is not a significant overhead.

\textbf{Future directions.} A potential future direction is replacing DeepLabv3+ with another state-of-the-art segmentation network and analyzing the impact. We expect to obtain similar results as we have already shown by analyzing the impact of backbone change in Section~\ref{sec:ablation}. So far, we have also understood that slightly different map aggregation policies can be beneficial in different architectures. As an example, one could replace the \texttt{Flush} operation that cleans up all the previous aggregated maps with \texttt{Retain} for which always a certain number of previous peaks (from previous cycles) are retained. In this work, we do neither claim nor prove a generic applicability of LoAd beyond semantic segmentation. Therefore, a more general understanding of the impact of LoAd in adversarial settings requires it to be applied to other applications which is outside the scope of this work.
\section*{Acknowledgment}
\vspace{-0.4cm}
The authors are thankful to Ahmad Zamanian for helpful discussions on adversarial networks, and to Saptarshi Das and Daniel Jeavons, and other members of Shell Global Solutions International B.V. for their support. The authors extend their appreciation to Ahmad Beirami from Facebook AI for pointers to DAGGER \cite{ross2011reduction} and sharing insights on the results reported in \cite{nouiehed2019solving, ostrovskii2020efficient}.
\newpage
{\small
	\bibliographystyle{ieeetr}
	\bibliography{LoAd_arxiv_v4}
}

\newpage
\appendix

\section{Supplementary Material}
\label{sec:recap}
\vspace{-0.4cm}
In the following, after a brief note on the limitations of pixel-wise cross-entropy (CE) loss, we dive deep into the mechanics of LoAd providing an extended and more elaborate version of Algorithm~\ref{alg:lookahead_simplified} besides a process flowchart view of the proposed method.
\section{On the Dimensionality of the Problem}
\label{sec:dim}
\vspace{-0.4cm}
Let us briefly reflect on why employing only a pixel-wise cross-entropy (CE) optimization loss for semantic segmentation and relying on gradual increase in receptive field of CNNs might not be sufficient for explicitly promoting a large set of candidate solutions for label maps. A good portion of these solutions might not be valid, bearing in mind that missing out on the rest could also lead to sub-optimal solutions for the segmentation problem. This is in line with literature \cite{liu2017deep, chen2017deeplab, shen2017semantic, liu2017deep, ke2018adaptive, zhao2019region, zhao2019correlation} proposing different approaches to address this challenge. Following our notation in Section~\ref{sec:adversarial}, let $\mathcal{D}_t =\{({\bf X},{\bf Y})_{1},...,({\bf X},{\bf Y})_{M}\}$ be training dataset with $M$ samples so that for all $({\bf X},{\bf Y})_{i} \in \mathcal{D}_t, ({\bf X},{\bf Y})_{i} \sim P({\bf X},{\bf Y})$ where $\mathcal{X}_t=\{{\bf X}|({\bf X},{\bf Y}) \in \mathcal{D}_t \}$ and $\mathcal{Y}_t=\{{\bf Y}|({\bf X},{\bf Y}) \in \mathcal{D}_t \}$ respectively denote the set of images and their corresponding label maps in the training dataset. As discussed earlier, ${\bf X}$ is of size $H \times W \times 3$ for RGB images with a total of $N$ pixels and the corresponding label map ${\bf Y}$ is of size $H \times W$ with elements in $\mathcal{K}=\{1,\dotsc,K\}$ where $K$ is the number of classes in segmentation task. Let $P({\bf X})$, $P({\bf Y})$ and $P({\bf X}, {\bf Y})$ denote probability mass functions of corresponding categorical distributions. Now, with pixel independence assumption:
\begin{equation} \label{independenceAssumptionEquation}
P({\bf Y})=\prod_{i=1}^{N}  P(y_{i}), 
\end{equation}
where $y_{i}$ denotes the $i$th pixel in ${\bf Y}$. This obviously ignores the potential correlation among the pixels. $P({\bf Y})$ is a function from the set $E_{\bf Y}$, comprised of all possible values ${\bf Y}$ can take, to a probability in $\mathbb{R}$. As such, learning $P({\bf Y})$ mandates exploring $E_{\bf Y}\times\mathbb{R}$. But how do all possible maps in $E_{\bf Y}$ look like? To answer this, let us consider a toy setup where ${\bf Y}$ is a $2\times1$ label map $\left[\protect\begin{smallmatrix}\ y_{1} & \\ \  y_{2}\protect\end{smallmatrix}\right]$ with $K=3$ classes in $\mathcal{K} = \{1, 2, 3\}$. The set of all possible label maps is simply: 
%
\begin{align*}
\notag
E_{\bf Y}=\left\{
\begin{bmatrix}
1 \\
1 \\
\end{bmatrix},
\begin{bmatrix}
2 \\
1 \\
\end{bmatrix},
\begin{bmatrix}
3 \\
1 \\
\end{bmatrix},
\begin{bmatrix}
1 \\
2 \\
\end{bmatrix},
\begin{bmatrix}
2 \\
2 \\
\end{bmatrix},
\begin{bmatrix}
3 \\
2 \\
\end{bmatrix},
\begin{bmatrix}
1 \\
3 \\
\end{bmatrix},
\begin{bmatrix}
2 \\
3 \\
\end{bmatrix},
\begin{bmatrix}
3 \\
3 \\
\end{bmatrix}
\right\},
\end{align*}
whose size is equal to the permutations with replacement of 2 pixels from a set of $3$ classes, i.e., $3^{2}=9$. One can straightforwardly generalize this toy example to any arbitrary number of pixels $N$ and classes $K$ and show that $|E_{\bf Y}|= K^{N}$ where $|\mathcal{A}|$ denotes the cardinality of set $\mathcal{A}$. Following the same approach, this time with pixel independence assumption in \eqref{independenceAssumptionEquation} for $y_1$ and $y_2$, the set of permutations would be $6$, counting $3$ possible values per $y_i$ summing them up due to independence. Again, it is straightforward to show $|E_{{\bf Y}}|= KN$. This toy example illustrates how the solution space of the problem with pixel independence assumption is significantly smaller than the one considering all possible correlations between pixels. However, the latter can easily become intractable for large size images, and that is the reason why a pixel-wise loss has been so popular. We think a conditional adversarial approach has the capacity to partially explore the larger solution space and to capture these pixel correlations in a general (and not only local) fashion. 


\section{Deeper Dive in Lookahead Adversarial Learning (LoAd)}
\label{sec:dive}
\begin{figure}[!ht]
	\centering
	\includegraphics[width=.85\textwidth]{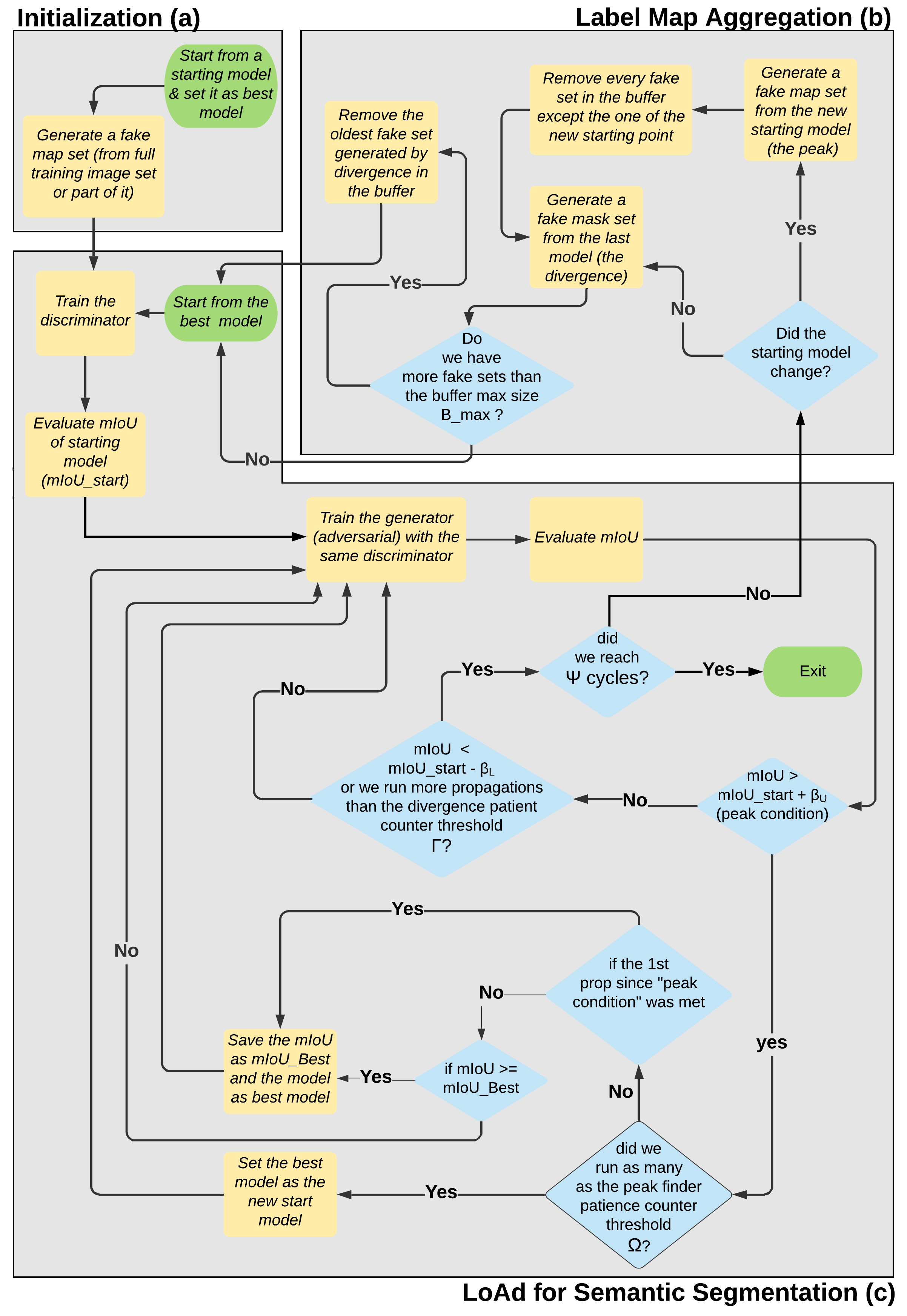}
	\vspace{-0.4cm}
	\caption{Process flow of LoAd with label map aggregation for semantic segmentation.}\label{fig:flowchart}
\end{figure}

\vspace{-0.4cm}
As discussed in Section~\ref{sec:lookaeahd}, there are subtle details we skipped or described on a high level to convey the core idea of the proposed method (LoAd) and not to steal the readers attention away from the main message. In this section, we dive deeper into the mechanics of LoAd using Fig.~\ref{fig:flowchart}, a flow chart that breaks down its process flow in clear terms, together with a more detailed version of Algorithm~\ref{alg:lookahead_simplified}. For the sake of clarity, we are using line numbers in Algorithm~\ref{alg:lookahead_detailed} to be able to refer to specific lines throughout the following explanations. Except for more details, the most important difference between Algorithm~\ref{alg:lookahead_detailed} and its simplified version, Algorithm~\ref{alg:lookahead_simplified}, is that the search process to find a new peak per cycle is conducted in a dynamic fashion by introducing a new parameter, $\omega$, that we call peak finder patience counter. The flowchart in Fig.~\ref{fig:flowchart} does not go down into the smallest details of every single line in Algorithm~\ref{alg:lookahead_detailed} yet it reflects on the most important steps to clarify the process. As can be seen in Fig.~\ref{fig:flowchart}, there are three components: a) initialization (a one-off process), b) label map aggregation (corresponding to Algorithm~\ref{alg:buffer}), and c) the main body of LoAd for semantic segmentation (corresponding to Algorithm~\ref{alg:lookahead_detailed}). The process starts from initialization depicted on the top left (part a), which corresponds to the preamble of Algorithm~\ref{alg:lookahead_detailed}. Provided a starting model, the cycles of LoAd can start and iterate between its main body (part c) and its label map aggregation module (part b).
\LinesNumbered
\begin{algorithm}[!ht]
	\SetKwInOut{Init}{Initialize}
	\SetAlgoLined
	\DontPrintSemicolon
	\SetNoFillComment
	\Init{$\psi = 0$, $g^{s} = g_{0}$, $\mathcal{B} = g^{s}(\mathcal{X}_t)$}
	\KwIn{\textup{maximum cycle:}\,$\Psi$, \textup{maximum divergence:}\,$\Gamma$, \textup{maximum peak finder:}\,$\Omega$, $\beta_l$, $\beta_u$}
	$\mu^{s}, \mu, \mu^{*} \gets \texttt{\textup{Evaluate mIoU}}$ \;
	\texttt{\textup{Train Discriminator}}$(\mathcal{D}_t \cup \mathcal{B})$\;
	\While{$\psi < \Psi$}{ 
		keep the very first starting point: $\mu^{0} \gets \mu^{s}$ \;
		\textup{start a divergence patience counter:} $\gamma \gets 0$\;
		\textup{start a peak finder patience counter:} $\omega \gets 0$\;
		\While{$\mu^{s} - \beta_l  < \mu$ \textup{and} $\gamma < \Gamma$ }{
			\textup{update the model:} $g \gets \texttt{\textup{Train Adversarial}}$\;
			$\mu \gets \texttt{\textup{Evaluate mIoU}}$ \;
			\If{$\mu > \mu^{s} + \beta_u$}{
				\If{\textup{new peak is found:} $\mu > \mu^{*}$}{
					\textup{update best performance:} $\mu^{*} \gets \mu$ \;
					\textup{update best model:} $g^{*} \gets g$ \;
				}
				\If{$\omega > \Omega$}{
					\text{update starting performance:} $\mu^{s} \gets \mu$\;
					\text{update starting model:} $g^{s} \gets g$ \;
					\textup{reset peak finder counter:} $\omega \gets 0$\; 
					\textup{reset divergence counter:} $\gamma \gets 0$ 
					
				}
				$\omega \gets \omega + 1$\;
			}
			$\gamma \gets \gamma + 1$\;
		}
		$g^{e} \gets \textup{keep last model of the cycle}$\;
		\eIf{\textup{best peak is better than very first start:} $\mu^{*} > \mu^{0}$}{
			\textup{set best model as start model:} $g^{s} \gets g^{*}$\;
			\textup{set best mIoU as start mIoU:} $\mu^{s} \gets \mu^{*}$\;
			$\mathcal{B} \gets$ \texttt{\textup{Lookahead Map Aggregation}}$(g^{*}, g^{e}, \mathcal{X})$ \;
			\textup{reset cycle counter} $\psi \gets 0$\;
		}{
			$\mathcal{B} \gets$ \texttt{\textup{Lookahead Map Aggregation}}$(0, g^{e}, \mathcal{X})$ \;
			\textup{start a new cycle} $\psi \gets \psi + 1$\;
		}
		\texttt{\textup{Train Discriminator}}$(\mathcal{D}_t \cup \mathcal{B})$\;
	}
	\caption{LoAd for Semantic Segmentation - Detailed Version}\label{alg:lookahead_detailed}
\end{algorithm}
Let us delve deeper into this process by zooming into Algorithm~\ref{alg:lookahead_detailed}. Lines $3$ to $35$ outline what happens in a single cycle of LoAd. A cycle starts by initializing the divergence patience counter $\gamma \in [0, \cdots, \Gamma]$ and the peak finder patience counter $\omega \in [0, \cdots, \Omega]$. At the beginning of every cycle, we mark the very initial model performance (in mIoU) as $\mu^{0}$ which will only be used in line $26$ to decide if throughout the cycle we actually found a new model (a new peak in the convergence graph) that is better than the very initial model of the cycle in mIoU sense. Lines $7$ to $24$ embody the main while loop of the cycle that keeps training the adversarial network until one of the following conditions is met: 1) the current mIoU ($\mu$) touches the bottom line we defined for downward/divergence trend ($\mu^{s} - \beta_l  < \mu$) or $\gamma$ meets its limit in the number of iterations. Inside this while loop, we watch for finding a new peak (model) in lines $10$ to $22$, and such a peak is only valid if the current mIoU goes beyond the starting mIoU plus a minimum increment ($\mu > \mu^{s} + \beta_u$). We came to understand by extensive experimentation that if the starting model ($g^s$ with mIoU $\mu^s$) is kept unchanged in this while loop (e.g., kept as $\mu^{0}$ as in line $4$), and thus, we select a peak based on a static peak finder patience counter $\omega$, we can prematurely kill an upward trend leading to a higher peak. This is the rationale behind the dynamic peak finding process in which we dynamically update the starting model ($g^s$, $\mu^s$) in lines $16$ and $17$. To do so, every time we reach the end of a peak finding process ($\omega$ reaches $\Omega$), if the mIoU is beyond the current starting model mIoU by at least $\beta_u$ in mIoU sense, we reset both $\gamma$ and $\omega$ and allow this upward trend to continue and keep updating the best model. This is delineated in lines $15$ to $20$. 

At the end of this inner while loop (lines $7$ to $24$), we are expected to mark one or two models: an ending model and possibly a new peak. The ending model of the cycle $g^{e}$ (line $25$) will be passed to the label map aggregation module regardless of whether we find a new peak or not. In case the dynamic peak finding process discovers a new peak that is better than the initial model of the cycle ($\mu^{*} > \mu^{0}$), then we pass both the new peak model $g^{*}$ and the ending model $g^{e}$ to our label map aggregation module. Lines $29$ and $32$ highlight the two ways in which the label map aggregation module can be invoked, with and without a new peak found. Based on that the map aggregation buffer $\mathcal{B}$ will be updated at the end of each cycle. The containment of the updated buffer will be concatenated with the full training dataset ($\mathcal{D}_t \cup \mathcal{B}$) and will be used to retrain the discriminator (line $35$). As discussed in Section~\ref{sec:adversarial}, we do not follow a standard GDA approach and every time train the discriminator with this dynamically updated dataset until a sufficient accuracy is reached. Lastly, a new peak serves as a new starting model based on which counting the cycles will be restarted (lines $27$, $28$ and $30$). On the other hand, if no new peak is found, we go back to where we started and continue (lines $32$ and $33$) hoping for finding a new peak in the next cycle. We fully stop the algorithm if we run $\Psi$ cycles and a new peak is not found. This process is also sketched in Fig.~\ref{fig:flowchart}.   
\newpage

\end{document}